\documentclass[11pt]{article}
\usepackage{acl_under_review}
\usepackage{times}
\usepackage{latexsym}
\usepackage[T1]{fontenc}
\usepackage[utf8]{inputenc}
\usepackage{microtype}
\usepackage{amsmath,amssymb,amsthm,mathtools}
\usepackage{booktabs,array}
\usepackage{graphicx}
\usepackage{placeins}
\usepackage{pgfplots}
\pgfplotsset{compat=1.18}
\usepackage{xcolor}
\usetikzlibrary{arrows.meta,positioning}
\usepackage{hyperref}

\newcommand{\R}{\mathbb{R}}
\newcommand{\E}{\mathbb{E}}
\newcommand{\norm}[1]{\lVert #1\rVert}
\newcommand{\ip}[2]{\langle #1,#2\rangle}
\newcommand{\calM}{\mathcal{M}}
\newcommand{\calC}{\mathcal{C}}
\newcommand{\calD}{\mathcal{D}}
\newcommand{\calI}{\mathcal{I}}

\newcommand{\newnum}[1]{#1}
\newtheorem{proposition}{Proposition}

\title{From Spectral Saliency to Finite Deletion: Evaluating Selective Language Model Edits}
\author{Ibne Farabi Shihab\thanks{Equal contribution.}\textsuperscript{,}\thanks{Corresponding author: \href{mailto:ishihab@iastate.edu}{ishihab@iastate.edu}.} \\
Department of Computer Science \\
Iowa State University, USA \\
\And
Joyanta Jyoti Mondal\footnotemark[1] \\
Department of Computer and \\Information Sciences \\
University of Delaware, USA\\
}
\date{}

\begin{document}
\maketitle

\begin{abstract}
Deleting singular components offers a precise way to remove a capability from a language model, yet the first-order spectral saliency used to rank components predicts only a local loss change, whereas deletion is a finite intervention. In this paper, we study how prediction error affects selective editing, where selective editing requires degradation of a target task while preserving others. We decompose the error into derivative transfer across item splits and nonlinear response along the deletion path. We establish that outer interval screening retains every feasible candidate when every candidate's intervals contain its finite validation loss changes on every task. We evaluate these effects using one answer-only negative log-likelihood objective for search, validation, and test, with disjoint calibration and evaluation masks. The primary configuration targets arithmetic in Mistral-7B-Instruct-v0.3 while preserving code and question answering; extensions cover Llama-3.1-8B-Instruct and reversed target roles. In the primary study, which deletes 1024 coordinates from a support of 7168, calibrated intervals cover 968 of 1024 evaluation masks. Outer screening reaches the direct-selected edit after 25 verification checks, whereas direct verification abstains at a 32-check cap, but it rejects 4 of 74 feasible masks. All three selected edits pass validation, yet only one meets every test constraint. Interval coverage, feasible-candidate retention, and independent test preservation are therefore distinct properties and require separate assessment.
\end{abstract}

\section{Introduction}

Selective editing degrades a target task while preserving others, a goal shared with language-model unlearning \citep{yao2024llmunlearning,li2024wmdp,liu2025rethinking}, and first-order spectral saliency ranks candidate edits by predicting how the loss changes near the unedited parameters. Deleting an entire singular component, however, moves the parameters by a finite amount, so predicted and measured loss changes can disagree. The disagreement matters for selection: it can discard a feasible edit or send an infeasible one to verification. We study how prediction error varies with deletion energy and how it changes candidate selection.

\begin{figure*}[t]
\centering
\includegraphics[width=\textwidth]{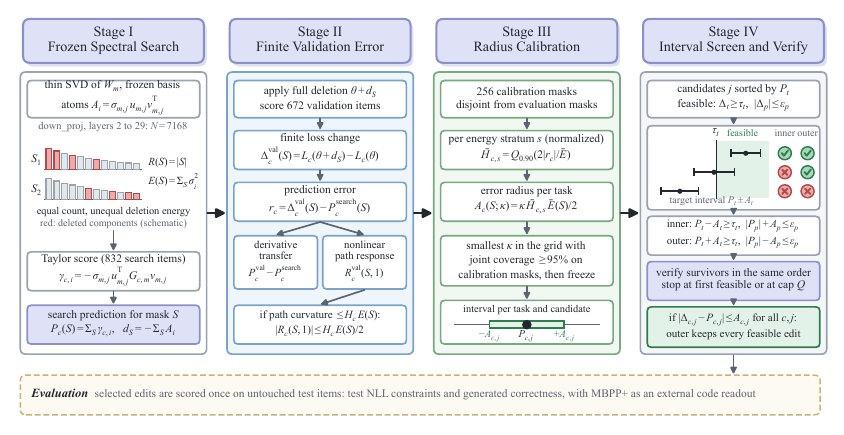}
\caption{Finite spectral deletion and interval screening. Stage I forms search predictions in a frozen singular basis. Stage II measures finite validation error and separates derivative transfer from nonlinear response along the deletion path. Stage III calibrates energy-normalized radii on masks disjoint from the evaluation pool. Stage IV screens candidates and verifies survivors in the fixed order under a check cap. The development pilot precedes the final freeze; Proposition~\ref{prop:outer-retention} states the retention condition.}
\label{fig:measurement-flow}
\end{figure*}

We define edits in a frozen singular value decomposition (SVD) basis. Each coordinate is a rank-one component, and each candidate deletes a subset of components. The coordinate count is the number of deleted components; the deletion energy is the squared size of the parameter displacement. Subsets with the same count can remove different amounts of energy, so we evaluate prediction error against energy at a fixed coordinate budget.

We use one end-to-end answer-only negative log-likelihood (NLL) objective for search gradients, validation scoring, calibration, and selected-edit test losses. NLL uses the same token and item normalization at every stage. Two differences remain: search and validation use different examples, and reference likelihood measures a different outcome from generated correctness. Derivative-transfer measurements and separate correctness readouts address them.

Figure~\ref{fig:measurement-flow} summarizes the method. Search derivatives predict the loss change of a fixed deletion, and validation scoring measures the finite change. We relate their difference to deletion energy, fit error radii on separate calibration masks, and measure coverage and pruning on evaluation masks. Test evaluation then scores selected edits on untouched items.

The primary study targets arithmetic in Mistral-7B-Instruct-v0.3. Each mask deletes 1024 coordinates from a support of 7168, and the study uses 1024 evaluation masks and 256 calibration masks. Before these pools are frozen, a separate 64-mask development pilot checks that the support and thresholds produce an informative range of candidates. Pilot outcomes may guide configuration changes; final evaluation outcomes never guide mask selection.

We compare three candidate-selection policies that share the same predictions. Direct verification checks candidates in target-prediction order. Inner screening keeps a candidate only when its entire predicted error rectangle satisfies the constraints. Outer screening keeps it whenever that rectangle contains a loss vector that satisfies the constraints, then verifies the survivors. Valid intervals guarantee that outer screening retains feasible edits, whereas calibrated intervals require a measured retention audit. Under a common check cap, pruning determines how far each policy progresses through the candidate order.

Our contributions are:
\begin{itemize}\setlength{\itemsep}{1pt}\setlength{\parsep}{0pt}\setlength{\topsep}{2pt}
\item \textbf{An error decomposition for finite spectral deletion.} The gap between a first-order search prediction and the finite validation loss change decomposes into derivative transfer across item splits and nonlinear response along the deletion path. A pathwise curvature condition bounds the latter term by deletion energy (Section~\ref{sec:setup}).
\item \textbf{A retention condition for interval screening.} When every candidate's intervals contain its finite validation changes on every task, outer screening retains every feasible candidate, and uncapped verification in the direct order returns the same first feasible output with no more checks (Proposition~\ref{prop:outer-retention}).
\item \textbf{An evaluation of error, retention, budgeted verification, and generated correctness under a fixed protocol.} On Mistral-7B-Instruct-v0.3, calibrated intervals jointly cover \newnum{968} of \newnum{1024} evaluation masks, and outer screening retains \newnum{70} of \newnum{74} feasible candidates. It reaches the direct-selected edit in \newnum{25} checks, whereas direct verification abstains at a \newnum{32}-check cap. All three selected edits pass validation, but only one meets the test constraints. The study extends to Llama-3.1-8B-Instruct and reversed target roles (Sections~\ref{sec:verification} and \ref{sec:readouts}).
\end{itemize}

\section{Related Work}

\paragraph{Curvature-based pruning.} Optimal Brain Damage uses diagonal second derivatives; Optimal Brain Surgeon uses inverse-Hessian information and compensates deletion by updating retained weights \citep{lecun1989obd,hassibi1992obs}. SparseGPT scales reconstruction updates to large language models (LLMs) through a quadratic layer-output objective \citep{frantar2023sparsegpt}. Wanda scores weight magnitude multiplied by input-activation norms, without weight reconstruction \citep{sun2024wanda}. Kronecker-factored approximate curvature (K-FAC) approximates Fisher blocks through Kronecker factors \citep{martens2015kfac}, and LLM Surgeon applies these approximations to target-loss pruning with correlated updates \citep{vanderouderaa2024llmsurgeon}. Our frozen spectral pool does not evaluate these methods' native compensated interventions or search procedures. Positive-semidefinite Fisher or reconstruction surrogates also differ from the signed end-to-end task Hessian analyzed here.

\paragraph{Taylor scores and spectral edits.} Taylor saliency estimates local loss effects \citep{molchanov2017pruning,molchanov2019importance}, including in structural LLM pruning \citep{ma2023llmpruner}; influence functions make similar local predictions and can be fragile in deep networks \citep{koh2017influence,basu2021fragile}. Weighted, activation-aware, and truncation-aware SVD methods optimize low-rank compression \citep{hsu2022fwsvd,yuan2023asvd,wang2025svdllm,wang2025svdllmv2}; SVD-Surgeon studies singular-value saliency and second-order compensation \citep{safari2026svdsurgeon}. LASER selects finite layerwise tail truncations using validation accuracy \citep{sharma2024laser}. We use fixed spectral edits to evaluate search-to-validation prediction error and its effect on verification; the contribution is this analysis and comparison, not a new first-order saliency formula or a reproduction of those complete pruning pipelines.

\paragraph{Editing and capability preservation.} ROME and MEMIT edit factual associations \citep{meng2022locating,meng2023mass}; localization alone need not identify the most effective intervention \citep{hase2023localization}. Task arithmetic and its linearized variant \citep{ilharco2023task,ortizjimenez2023tangent}, unlearning by gradient ascent, preference objectives, or pruning \citep{jang2023knowledge,eldan2023harry,zhang2024npo,pochinkov2024dissecting,li2024wmdp}, and representation erasure or direction ablation \citep{ravfogel2020null,belrose2023leace,arditi2024refusal} motivate targeted changes with explicit preserved-task readouts, as in unlearning benchmarks \citep{maini2024tofu,shi2025muse}. Pruning and low-rank weight edits can isolate safety-critical regions \citep{wei2024brittleness}, and relearning or adversarial probes can recover unlearned knowledge \citep{hu2025jogging,lucki2025adversarial}; we do not evaluate such recovery. Our reference-NLL constraints measure a narrower form of preservation than an underlying capability, so we report generated correctness and selection abstention separately.

\section{Method: Interval Screening for Finite Spectral Deletion}
\label{sec:setup}

This section defines the intervention, constraints, prediction error, and error radii, then states the screening rules. Table~\ref{tab:notation} collects the notation.

\subsection{Problem Formulation}
\label{sec:problem}

Let $f_\theta$ be a decoder-only language model and $\calM$ a collection of editable matrices. For a linear-layer weight $W_m\in\R^{d_m^{\rm out}\times d_m^{\rm in}}$, the row-vector forward map is $x\mapsto xW_m^\top$, and its thin SVD is
\begin{equation}
\begin{aligned}
 W_m&=\sum_{j=1}^{r_m}\sigma_{m,j}u_{m,j}v_{m,j}^{\top},\\
 r_m&=\min(d_m^{\rm out},d_m^{\rm in}),
\end{aligned}
\end{equation}
with nonincreasing nonnegative singular values. We fix the searchable bands $\mathcal J_m$ and collect their coordinates in $\calI=\{(m,j):m\in\calM,\ j\in\mathcal J_m\}$, with $N=|\calI|$. For $i=(m,j)$, the spectral atom $A_i=\sigma_{m,j}u_{m,j}v_{m,j}^{\top}$ occupies matrix $m$ and is zero elsewhere. Selecting $S\subseteq\calI$ gives the deletion direction $d_S=-\sum_{i\in S}A_i$. The amplitude $\alpha\in[0,1]$ controls the edit size: parameters become $\theta+\alpha d_S$, and full deletion corresponds to $\alpha=1$.

The coordinate count $R(S)=|S|$ pools selected components across the network, although each matrix update has its own rank. The deletion energy is
\begin{equation}
 E(S)=\norm{d_S}_F^2=\sum_{i\in S}e_i,
 \qquad e_i=\sigma_{m,j}^2.
 \label{eq:energy}
\end{equation}
Within matrix $m$, the update rank is the number of selected positive singular values and its operator norm is the largest selected singular value; both follow from singular-component orthogonality. When singular values repeat, deleting an individual component depends on the chosen basis, whereas deleting the entire repeated-value subspace does not. Appendix~\ref{app:implementation} describes the implementation of the basis and edits.

For capability $c$, fix an evaluation distribution $\calD_c$ and differentiable loss
\begin{equation}
 L_c(\theta)=\E_{(x,y)\sim\calD_c}[\ell_c(f_\theta(x),y)].
\end{equation}
We define the loss change as $\Delta_c(S,\alpha)=L_c(\theta+\alpha d_S)-L_c(\theta)$, with positive values indicating degradation. For the separate higher-is-better score $\operatorname{Acc}_c$, we define the score drop $D_c(S,\alpha)=\operatorname{Acc}_c(\theta)-\operatorname{Acc}_c(\theta+\alpha d_S)$. For the NLL objective used here, loss change measures reference likelihood and accuracy drop measures correctness.

We fix a task-specific target floor $\tau_t>0$, preserved tolerances $\epsilon_p\geq0$, an integer $B\geq0$, and $E_{\max}\geq0$. An edit is selective at these thresholds if
\begin{equation}
\begin{aligned}
 R(S)&\leq B, & E(S)&\leq E_{\max},\\
 \Delta_t(S,1)&\geq\tau_t, & |\Delta_p(S,1)|&\leq\epsilon_p\quad(p\in P),
\end{aligned}
\label{eq:selective}
\end{equation}
Here $t$ is the target and $P\subseteq\calC\setminus\{t\}$ contains the preserved tasks. Absolute preservation requires stable preserved losses and therefore penalizes both improvement and deterioration beyond tolerance. Harm-only preservation instead requires $\Delta_p(S,1)\leq\epsilon_p$ and permits improvements of any size. Generated correctness is evaluated separately under either criterion. Shared token-normalized units do not make a target floor or tolerance equally meaningful across tasks.

\subsection{Spectral Saliency Prediction}

For $G_{c,m}=\nabla_{W_m}L_c(\theta)$, the spectral Taylor score is
\begin{equation}
 \gamma_{c,i}=-\sigma_{m,j}u_{m,j}^{\top}G_{c,m}v_{m,j}.
 \label{eq:gamma}
\end{equation}
In the frozen SVD basis, this score is the derivative with respect to a component's deletion amplitude. Its sign predicts whether a small deletion raises or lowers the task loss, a direction the singular value alone does not determine. The search prediction of a mask sums these scores on the search split, $P_c^{\rm search}(S)=\sum_{i\in S}\gamma_{c,i}^{\rm search}$.

\subsection{Finite Deletion Error}

For $g_c(\alpha)=L_c(\theta+\alpha d_S)$, twice continuous differentiability along the edit path gives the exact identity
\begin{equation}
\begin{aligned}
 \Delta_c(S,\alpha)&=\alpha\sum_{i\in S}\gamma_{c,i}+\mathcal R_c(S,\alpha),\\
 \mathcal R_c(S,\alpha)&=\int_0^\alpha(\alpha-u)h_{c,S}(u)\,du,\\
 h_{c,S}(u)&=d_S^\top\nabla^2L_c(\theta+u d_S)d_S.
\end{aligned}
\label{eq:taylor}
\end{equation}
If $|h_{c,S}(u)|\leq H_c E(S)$ along that path, then
\begin{equation}
 |\mathcal R_c(S,\alpha)|\leq\tfrac12\alpha^2H_cE(S).
 \label{eq:remainder}
\end{equation}
The bound depends on deletion energy and on a pathwise curvature condition; a coordinate budget alone gives no such guarantee. A large residual shows that the finite response departs from the local linear prediction, but not that the Hessian at $\alpha=0$ explains the whole discrepancy. Appendix~\ref{app:measured-path} defines the measured test-split residual.

The observable validation error is
\begin{equation}
\begin{aligned}
 r_c^{\rm val}(S)&=\Delta_c^{\rm val}(S,1)-P_c^{\rm search}(S),\\
 &=\bigl(P_c^{\rm val}-P_c^{\rm search}\bigr)+\mathcal R_c^{\rm val}(S,1).
\end{aligned}
\label{eq:validation-error}
\end{equation}
Equation~\ref{eq:validation-error} separates two sources of error: the derivative changes between search and validation items, and the finite deletion produces a nonlinear response on validation. Forward scoring of the full pool measures their sum, which is the error the screening radius must cover; recomputing validation derivatives on selected frozen masks separates the two terms. A Hessian measured only at the unedited model bounds neither their sum nor the response over the full path.

To compare masks within a searchable band, we normalize energy as $\bar E=E/E_{\rm band}$ and curvature as $\bar H=HE_{\rm band}$. Since $\bar H\bar E=HE$, this rescales without changing the Taylor allowance. The normalization is specific to the checkpoint and support and does not make coefficients transferable between checkpoints. Appendix~\ref{app:curvature-proof} gives the decomposition of base-point and path curvature and the conditions for its bounds.

\subsection{Calibrated Error Radii}
\label{sec:radii}

Error radii are fit on calibration masks, disjoint from evaluation masks, analogous to split conformal calibration \citep{lei2018distribution}; for those in deletion-energy stratum $s$, we fit
\begin{equation}
\begin{aligned}
 \bar H_{c,s}&=Q_{0.90}
 \left(\frac{2|r_c^{\rm val}(S)|}{\bar E(S)}\right),\\
 A_c(S;\kappa)&=\tfrac12\kappa\bar H_{c,s}\bar E(S).
\end{aligned}
\label{eq:calibration}
\end{equation}
The quantile is the smallest sorted observation reaching the requested level. We choose the smallest multiplier in the declared grid \{0,0.25,0.5,0.75,1,1.25,1.5,2,3,4\} whose intervals jointly cover all tasks on at least 95\% of calibration masks, and freeze the coefficients and multiplier before examining evaluation outcomes. Empty calibration strata, candidates outside calibrated energy ranges, and failure to find a qualifying multiplier all trigger direct verification.

The radius covers search-to-validation prediction error, including both derivative transfer and nonlinear response, so separate masks measure its coverage. After the freeze, complete-pool quantiles and scaling summaries describe the policy's behavior but cannot alter its screening coefficients.

\subsection{Screening and Verification Policies}
\label{sec:policies}

All policies follow the same decreasing search-target-prediction order, with support-ID ties resolved before filtering. Each checks finite validation losses until it selects a candidate, exhausts its eligible pool, or reaches its cap $Q\in\{\newnum{32},\newnum{128},\newnum{512}\}$. Because filtering preserves order, the comparison isolates the effect of pruning.

For candidate $j$, let $P_{c,j}$ be the search prediction and $A_{c,j}\geq0$ its proposed error radius. The inner and outer tests are respectively
\begin{equation}
\begin{aligned}
\mathrm{Inner}:\quad&P_{t,j}-A_{t,j}\geq\tau_t,\quad
 |P_{p,j}|+A_{p,j}\leq\epsilon_p,\\
\mathrm{Outer}:\quad&P_{t,j}+A_{t,j}\geq\tau_t,\quad
 |P_{p,j}|-A_{p,j}\leq\epsilon_p.
\end{aligned}
\label{eq:outer-screen}
\end{equation}
The conditions apply to every preserved task. Harm-only preservation replaces the inner preserved inequality by $P_{p,j}+A_{p,j}\leq\epsilon_p$ and the outer inequality by $P_{p,j}-A_{p,j}\leq\epsilon_p$.

\begin{proposition}[retention under valid intervals]
\label{prop:outer-retention}
For a frozen pool satisfying its support, rank, and energy constraints, suppose $|\Delta_{c,j}^{\rm val}-P_{c,j}|\leq A_{c,j}$ for every task and candidate. Outer screening retains every validation-feasible candidate. Verification in the unchanged direct order, without a binding cap, returns the same first feasible output or the same abstention with no more checks. If the only information is the product of the task intervals, every outer survivor admits a feasible assignment.
\end{proposition}

The proof follows from interval intersection: if a feasible loss value lies in an interval, that interval intersects the task's allowed range, and keeping survivors in their original order preserves the first feasible candidate (Appendix~\ref{app:outer-proof} gives the full argument and boundary cases). The guarantee requires simultaneous interval validity for all tasks and candidates, which calibration coverage assesses empirically. With a binding check cap, policies can reach different portions of the order and return different outputs.

\section{Experiments}
\label{sec:results}

After the setup, we report prediction error versus deletion energy, coverage and retention, budgeted verification, and test preservation.

\subsection{Experimental Setup}
\label{sec:study}

The primary checkpoint is \texttt{mistralai/}\allowbreak\texttt{Mistral-7B-}\allowbreak\texttt{Instruct-v0.3} \citep{jiang2023mistral,mistral2024modelcard}. The target task is GSM8K arithmetic \citep{cobbe2021gsm8k}; the preserved NLL tasks are HumanEval code \citep{chen2021humaneval} and ARC-C question answering \citep{clark2018arc}. MBPP+, the EvalPlus extension of MBPP \citep{austin2021mbpp,liu2023evalplus}, provides the primary external code-correctness readout, with HumanEval supplementary. Table~\ref{tab:study-splits} in Appendix~\ref{app:deferred-protocol} gives the fixed item allocation. Each validation mask is scored on 672 items: 512 arithmetic, 32 code, and 128 question answering.

The loss averages answer-token NLL within each item, then averages across items in the split. Prompt tokens are masked. Gradients and finite scoring share tokenization, answer boundaries, end-of-sequence (EOS) handling, and truncation. Generated correctness uses its own fixed prompt and decoding policy and measures a separate outcome.

We edit \texttt{mlp.down\_proj} weights in zero-based layers 2 through 29 and search the leading 256 singular components of each matrix. Official metadata gives shape $4096\times14336$, yielding 28 matrices and 7168 searchable coordinates \citep{mistral2024configuration,mistral2024weightindex}. This contiguous interior-layer support holds projection type fixed; we make no claim of optimality. A preflight checks the modules against the pinned checkpoint before basis construction (Appendix~\ref{app:implementation}), and mask generation uses the frozen manifest, serialized basis, search IDs, and gradient arrays. Item-averaged gradients from 832 search items give the coefficients in Equation~\ref{eq:gamma}; Appendix~\ref{app:deferred-protocol} gives basis and gradient-audit details.

Before the final freeze, a separate 64-mask development pilot checks that the configuration yields an informative range of candidates. Pilot masks are excluded from both final pools and all summaries, and the 256-mask calibration and 1024-mask evaluation manifests are sealed before either pool is scored (Appendix~\ref{app:dev-pilot}).

Proposals meet the exact coordinate budget and use search predictions only; finite loss changes never enter their scores. Appendix~\ref{app:real-generator} specifies the proposal families, energy repair, seeds, and exclusions. Calibration follows Section~\ref{sec:radii}.

\subsection{Prediction Error versus Deletion Energy}
\label{sec:error-energy}

We summarize prediction error by task and deletion-energy stratum over \newnum{1280} final masks, keeping calibration and evaluation records separate and excluding development-pilot masks. Each record contains $P_c^{\rm search}$, $\Delta_c^{\rm val}$, signed $r_c^{\rm val}$, absolute error, energy, stratum, and support hash.

\begin{figure}[t]
\centering
\includegraphics[width=\columnwidth]{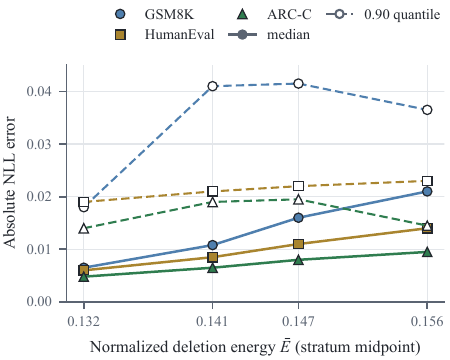}
\caption{Absolute NLL prediction error by deletion-energy stratum. Circles, squares, and triangles denote GSM8K, HumanEval, and ARC-C. Solid lines with filled markers show medians; dashed lines with hollow markers show \newnum{0.90} quantiles. Points use interval midpoints rather than median mask energies (Table~\ref{tab:error-quantiles}). Connecting lines guide comparison and are not fitted curves; Table~\ref{tab:scaling} reports the separate four-point aggregate fit.}
\label{fig:error-energy}
\end{figure}

Median absolute error increases across the energy strata (Figure~\ref{fig:error-energy}). Figure~\ref{fig:error-energy} plots the four stratum medians and \newnum{0.90} quantiles at energy-interval midpoints, and Table~\ref{tab:error-quantiles} (Appendix~\ref{app:error-quantiles}) adds the \newnum{0.95} quantiles and calibration-only coefficients for all three tasks. Absolute and normalized error answer different questions: the median error can rise while the fitted coefficient falls.

The energy trend is descriptive because proposal family and support composition vary with energy. Appendix~\ref{app:energy-scaling} gives the aggregate fit and optional component diagnostics; the frozen-pool trend does not establish a universal scaling law.

\subsection{Coverage and Feasible-Candidate Retention}
\label{sec:verification}

\begin{figure*}[t]
\centering
\includegraphics[width=\textwidth]{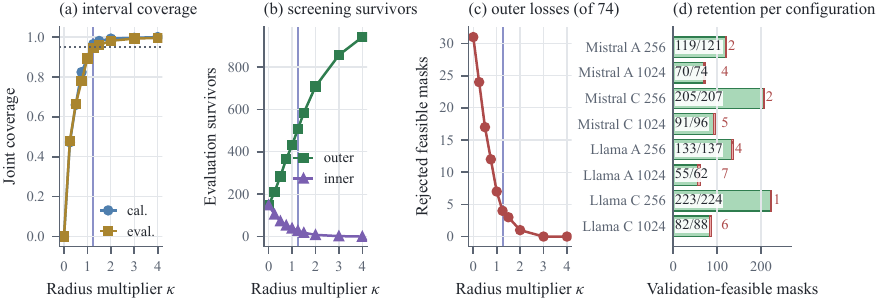}
\caption{Interval coverage and screening retention. (a) Joint coverage on \newnum{256} calibration and \newnum{1024} evaluation masks; dotted and vertical lines mark \newnum{95}\% coverage and the selected $\kappa=\newnum{1.25}$. (b) Inner and outer evaluation survivors. (c) Feasible masks rejected by outer screening, out of \newnum{74}. Panels (a) to (c) show the primary study. (d) Retained (green) and rejected (red, hatched) feasible masks at each configuration's selected multiplier. In-bar labels give retained/feasible counts; labels to the right give rejected counts. Labels identify checkpoint, target (A: arithmetic; C: code), and coordinate budget $B$.}
\label{fig:coverage-retention}
\end{figure*}

Coverage measures whether a candidate's intervals contain its finite validation loss changes on all tasks. Retention measures whether screening keeps a validation-feasible candidate. At the selected multiplier of \newnum{1.25}, the intervals cover \newnum{968} of \newnum{1024} evaluation masks, while outer screening retains \newnum{70} of \newnum{74} feasible masks. Figure~\ref{fig:coverage-retention} reports the multiplier sweep: calibration masks alone determine the multiplier, and the evaluation pool measures its coverage. At the selected multiplier, joint calibration coverage is \newnum{247} of \newnum{256} masks. Joint coverage is the intersection of the taskwise events; Table~\ref{tab:marginal-coverage} gives their marginal counts.

Calibrated outer screening rejects feasible masks even though it returns the direct-verification output. A feasible mask meets the finite validation target floor and both preserved tolerances; of the \newnum{74} such masks in the primary evaluation, outer screening retains \newnum{70} and rejects \newnum{4} (Figure~\ref{fig:coverage-retention}c).

Proposition~\ref{prop:outer-retention} guarantees retention only when all intervals are valid, so fitted intervals require measurements of both coverage and retention. An uncovered error either leaves the first feasible candidate intact or changes the selected output, and marginal error quantiles alone cannot distinguish these outcomes.

\subsection{Budgeted Verification}
\label{sec:budget-verification}


Pruning changes how far each policy progresses through the shared candidate order. The first direct-feasible candidate, A, occupies position \newnum{63} in the shared candidate order. Direct verification therefore requires \newnum{63} checks and abstains under a \newnum{32}-check cap. Calibrated outer screening removes earlier candidates and reaches A in \newnum{25} checks. Zero-radius screening rejects A because its code prediction exceeds tolerance and returns B after \newnum{17} checks, and calibrated inner verification reaches C after \newnum{7} checks. At the fitted multiplier, inner screening keeps \newnum{28} candidates and outer screening keeps \newnum{508}; both zero-radius filters keep \newnum{148}. The policies thus spend the same budget on different portions of the candidate order. Appendix~\ref{app:cross-verification} gives the check counts across checkpoints, targets, and budgets (Figure~\ref{fig:budget-checks}).

The calibrated outer screen rejects between \newnum{1} and \newnum{7} feasible masks in each configuration (Figure~\ref{fig:coverage-retention}d). At $Q=\newnum{128}$, it reaches the first direct-feasible output in all eight configurations, whereas direct verification exhausts the cap in Mistral/code/\newnum{1024}, Llama/arithmetic/\newnum{1024}, and Llama/code/\newnum{1024}. At $Q=\newnum{512}$, direct and outer verification agree on the selected support in every configuration. Filtering therefore changes access to the shared candidate order under a cap without guaranteeing retention of every feasible mask.

An abstention must be read through its cause: budget exhaustion, survivor exhaustion, calibration fallback, or an empty feasible pool; none establishes successful target removal. Every policy and budget selecting the same support reuses one endpoint record, so its reported losses and correctness stay consistent. Appendix~\ref{app:cost} accounts for the scoring workload; replayed consultation counts do not measure runtime.

\subsection{Test Preservation of Selected Edits}
\label{sec:readouts}

Test evaluation scores each distinct edit in the frozen union of supports selected across policies and check caps, plus the unedited baseline, once for NLL and correctness. GSM8K measures arithmetic answer correctness and ARC-C preserved question answering; MBPP+ is the primary external code readout and HumanEval is supplementary. A code completion is correct only if the fixed harness accepts it under the declared tests and timeout policy.

\begin{figure*}[t]
\centering
\includegraphics[width=\textwidth,height=2.35in,keepaspectratio]{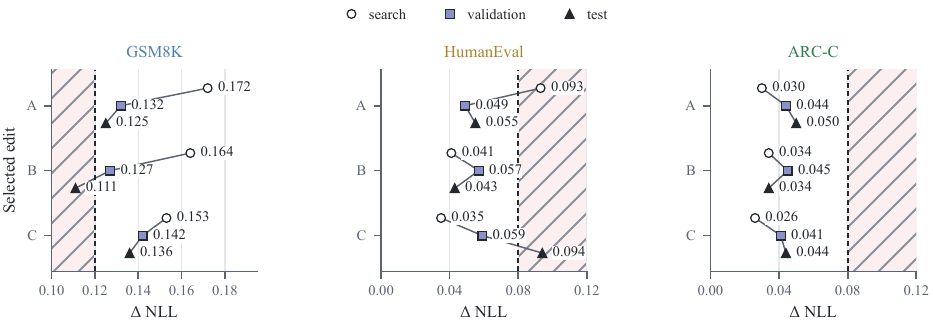}
\caption{Search predictions, finite validation changes, and test NLL changes for selected edits A, B, and C. Circles, squares, and triangles denote search, validation, and test. Hatched regions violate the \newnum{0.12} GSM8K target floor or the \newnum{0.08} HumanEval and ARC-C preservation tolerance. All edits pass validation; on test, B misses the target floor and C exceeds the code tolerance. Rows group each edit's search, validation, and test values; numeric labels give NLL changes to the reported three-decimal precision. All losses use the same answer-only objective on their respective splits.}
\label{fig:val-test}
\end{figure*}

\begin{figure*}[t]
\centering
\includegraphics[width=\textwidth]{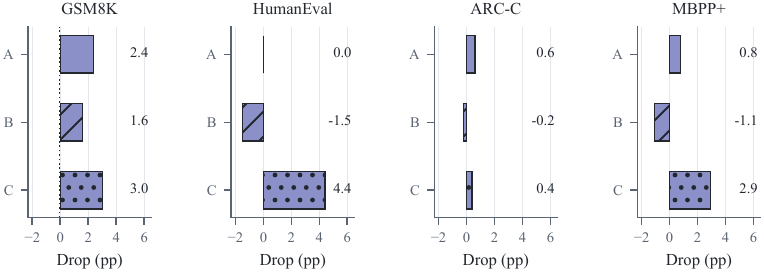}
\caption{Test correctness drops for primary selected edits A, B, and C. Positive values denote deterioration; negative values denote improvement. Bars use distinct hatches and report percentage-point drops. Denominators are \newnum{1319}, \newnum{68}, \newnum{1172}, and \newnum{378} for GSM8K, HumanEval, ARC-C, and MBPP+, respectively. Figure~\ref{fig:val-test} gives the corresponding calibrated-task NLL changes.}
\label{fig:primary-correctness}
\end{figure*}

Reference likelihood and correctness need not move together (Figures~\ref{fig:val-test} and \ref{fig:primary-correctness}). On MBPP+, A and C reduce correctness by \newnum{0.8} and \newnum{2.9} percentage points (pp), respectively, whereas B increases it by \newnum{1.1} pp. Positive NLL changes indicate worse reference likelihood, and positive accuracy drops indicate fewer correct outputs. Correctness drops use unrounded counts and are rounded only for display, so subtracting rounded accuracies can give a different value. An edit can exceed an NLL tolerance with little correctness change, or change correctness without a comparable shift in token-average loss. Quantifying uncertainty requires paired test-item resampling, and policies that select the same edit share one outcome record and are not independent experiments.

Validation feasibility does not ensure test feasibility. All three selected edits pass finite validation, but on test A meets the constraints, B misses the arithmetic floor, and C exceeds the code tolerance (Figure~\ref{fig:val-test}). Absolute preservation, which rejects improvements or deterioration beyond tolerance, and harm-only preservation, which rejects excessive positive loss changes, give the same decisions on these frozen outputs. Correctness remains a separate outcome even for edits that satisfy the NLL constraints.

The extension asks whether these behaviors recur across checkpoints and target roles. It crosses Mistral-7B-Instruct-v0.3 \citep{jiang2023mistral,mistral2024modelcard} and Llama-3.1-8B-Instruct \citep{dubey2024llama3,meta2024llama31instruct} with arithmetic and code targets at budgets \newnum{256} and \newnum{1024}, sharing the objective, generator rules, and accounting while fitting each checkpoint's own singular basis and calibration coefficients. Appendix~\ref{app:cross-model} reports coverage, retention, verification, and test outcomes for every configuration.

At $Q=\newnum{512}$, outer verification uses fewer checks than direct verification in every configuration (Figure~\ref{fig:budget-checks}). Reaching an edit in fewer checks does not ensure better test preservation; Appendix~\ref{app:cross-model} compares the selected outputs under each configuration's fixed thresholds.

\section{Discussion and Conclusion}

We analyze first-order prediction error for finite spectral deletions, separating derivative transfer across item splits from nonlinear response along the deletion path. When intervals contain every candidate's finite validation changes on every task, outer screening retains all feasible edits (Proposition~\ref{prop:outer-retention}). Calibrated intervals need not meet this premise: on Mistral-7B-Instruct-v0.3, outer screening reaches the direct-selected edit in \newnum{25} checks but rejects \newnum{4} of \newnum{74} feasible masks, and only one of three validation-feasible edits meets test constraints. Coverage, retention, budgeted selection, and test preservation therefore require separate assessment.

\section*{Limitations}

Our study covers two checkpoints, arithmetic and code targets, one proposal mechanism, and fixed task splits. It establishes neither optimality over unrestricted spectral subsets nor transfer of calibration coefficients to other checkpoints or supports; new checkpoints and target roles need their own bases and calibration.

Calibration and evaluation use disjoint masks but share the fixed validation corpus, which also serves the development pilot. Coverage estimates are therefore conditional on that corpus, the proposal mechanism, and the energy strata. Disjoint masks do not provide independent item samples. An empirical calibration quantile does not establish the uniform premise of Proposition~\ref{prop:outer-retention}; the evaluation audit therefore measures joint coverage and rejected feasible masks directly. The policy comparison uses one proposal order, and because deletion energy varies with support composition and proposal scores, the fitted energy relationship is descriptive.

Full-pool scoring adds audit cost, and cached replay counts do not establish runtime savings (Appendix~\ref{app:cost}). Reference likelihood and generated correctness measure different behavior; MBPP+ strengthens the external code readout, with HumanEval supplementary. Quantifying uncertainty and end-to-end utility requires paired item records, declared interval families, and measured timings.

\section*{AI Use Statement}
Generative AI tools assist with language editing, literature discovery, mathematical exposition, and source-code preparation. The authors verify the proofs, citations, experimental records, and final claims and remain responsible for the manuscript.

\section*{Reproducibility Statement}
The appendices give proofs of the theoretical results, the spectral intervention and capped verification procedure, and the controlled experimental settings. The experimental protocol specifies loss normalization, item splits, proposal generation, calibration, random seeds, and numerical comparison rules. The reported diagnostics and selected-edit results retain their stated numerical units across checkpoints, target roles, and spectral budgets.

\bibliography{custom}

@inproceedings{belrose2023leace,
  author = {Belrose, Nora and Schneider-Joseph, David and Ravfogel, Shauli and Cotterell, Ryan and Raff, Edward and Biderman, Stella},
  title = {{LEACE}: Perfect Linear Concept Erasure in Closed Form},
  booktitle = {Advances in Neural Information Processing Systems},
  volume = {36},
  year = {2023},
  url = {https://proceedings.neurips.cc/paper_files/paper/2023/hash/d066d21c619d0a78c5b557fa3291a8f4-Abstract-Conference.html}
}

@inproceedings{ilharco2023task,
  author = {Ilharco, Gabriel and Ribeiro, Marco Tulio and Wortsman, Mitchell and Gururangan, Suchin and Schmidt, Ludwig and Hajishirzi, Hannaneh and Farhadi, Ali},
  title = {Editing Models with Task Arithmetic},
  booktitle = {International Conference on Learning Representations},
  year = {2023},
  url = {https://arxiv.org/abs/2212.04089}
}

@inproceedings{meng2022locating,
  author = {Meng, Kevin and Bau, David and Andonian, Alex and Belinkov, Yonatan},
  title = {Locating and Editing Factual Associations in {GPT}},
  booktitle = {Advances in Neural Information Processing Systems},
  volume = {35},
  year = {2022},
  url = {https://proceedings.neurips.cc/paper_files/paper/2022/hash/6f1d43d5a82a37e89b0665b33bf3a182-Abstract-Conference.html}
}

@inproceedings{meng2023mass,
  author = {Meng, Kevin and Sharma, Arnab Sen and Andonian, Alex and Belinkov, Yonatan and Bau, David},
  title = {Mass-Editing Memory in a Transformer},
  booktitle = {International Conference on Learning Representations},
  year = {2023},
  url = {https://openreview.net/forum?id=MkbcAHIYgyS}
}

@inproceedings{ravfogel2020null,
  author = {Ravfogel, Shauli and Elazar, Yanai and Gonen, Hila and Twiton, Michael and Goldberg, Yoav},
  title = {Null It Out: Guarding Protected Attributes by Iterative Nullspace Projection},
  booktitle = {Proceedings of the 58th Annual Meeting of the Association for Computational Linguistics},
  year = {2020},
  pages = {7237--7256},
  publisher = {Association for Computational Linguistics},
  doi = {10.18653/v1/2020.acl-main.647},
  url = {https://aclanthology.org/2020.acl-main.647/}
}

@inproceedings{sharma2024laser,
  author = {Sharma, Pratyusha and Ash, Jordan T. and Misra, Dipendra},
  title = {The Truth Is in There: Improving Reasoning in Language Models with Layer-Selective Rank Reduction},
  booktitle = {International Conference on Learning Representations},
  year = {2024},
  url = {https://openreview.net/forum?id=ozX92bu8VA}
}

@inproceedings{molchanov2017pruning,
  author = {Molchanov, Pavlo and Tyree, Stephen and Karras, Tero and Aila, Timo and Kautz, Jan},
  title = {Pruning Convolutional Neural Networks for Resource Efficient Inference},
  booktitle = {International Conference on Learning Representations},
  year = {2017},
  url = {https://arxiv.org/abs/1611.06440}
}

@inproceedings{molchanov2019importance,
  author = {Molchanov, Pavlo and Mallya, Arun and Tyree, Stephen and Frosio, Iuri and Kautz, Jan},
  title = {Importance Estimation for Neural Network Pruning},
  booktitle = {Proceedings of the IEEE/CVF Conference on Computer Vision and Pattern Recognition},
  year = {2019},
  url = {https://openaccess.thecvf.com/content_CVPR_2019/html/Molchanov_Importance_Estimation_for_Neural_Network_Pruning_CVPR_2019_paper.html}
}

@inproceedings{ma2023llmpruner,
  author = {Ma, Xinyin and Fang, Gongfan and Wang, Xinchao},
  title = {{LLM-Pruner}: On the Structural Pruning of Large Language Models},
  booktitle = {Advances in Neural Information Processing Systems},
  volume = {36},
  year = {2023},
  url = {https://proceedings.neurips.cc/paper_files/paper/2023/hash/44956951349095f74492a5471128a7e0-Abstract-Conference.html}
}

@inproceedings{hase2023localization,
  author = {Hase, Peter and Bansal, Mohit and Kim, Been and Ghandeharioun, Asma},
  title = {Does Localization Inform Editing? Surprising Differences in Causality-Based Localization vs. Knowledge Editing in Language Models},
  booktitle = {Advances in Neural Information Processing Systems},
  volume = {36},
  year = {2023},
  url = {https://proceedings.neurips.cc/paper_files/paper/2023/hash/3927bbdcf0e8d1fa8aa23c26f358a281-Abstract-Conference.html}
}

@inproceedings{li2024wmdp,
  author = {Li, Nathaniel and Pan, Alexander and Gopal, Anjali and Yue, Summer and Berrios, Daniel and Gatti, Alice and Li, Justin D. and Dombrowski, Ann-Kathrin and Goel, Shashwat and Mukobi, Gabriel and Helm-Burger, Nathan and Lababidi, Rassin and Justen, Lennart and Liu, Andrew Bo and Chen, Michael and Barrass, Isabelle and Zhang, Oliver and Zhu, Xiaoyuan and Tamirisa, Rishub and Bharathi, Bhrugu and Herbert-Voss, Ariel and Breuer, Cort B. and Zou, Andy and Mazeika, Mantas and Wang, Zifan and Oswal, Palash and Lin, Weiran and Hunt, Adam Alfred and Tienken-Harder, Justin and Shih, Kevin Y. and Talley, Kemper and Guan, John and Steneker, Ian and Campbell, David and Jokubaitis, Brad and Basart, Steven and Fitz, Stephen and Kumaraguru, Ponnurangam and Karmakar, Kallol Krishna and Tupakula, Uday and Varadharajan, Vijay and Shoshitaishvili, Yan and Ba, Jimmy and Esvelt, Kevin M. and Wang, Alexandr and Hendrycks, Dan},
  title = {The {WMDP} Benchmark: Measuring and Reducing Malicious Use with Unlearning},
  booktitle = {Proceedings of the 41st International Conference on Machine Learning},
  year = {2024},
  volume = {235},
  series = {Proceedings of Machine Learning Research},
  pages = {28525--28550},
  publisher = {PMLR},
  url = {https://proceedings.mlr.press/v235/li24bc.html}
}

@inproceedings{wang2025svdllm,
  author = {Wang, Xin and Zheng, Yu and Wan, Zhongwei and Zhang, Mi},
  title = {{SVD-LLM}: Truncation-Aware Singular Value Decomposition for Large Language Model Compression},
  booktitle = {International Conference on Learning Representations},
  year = {2025},
  url = {https://openreview.net/forum?id=LNYIUouhdt}
}

@inproceedings{wang2025svdllmv2,
  author = {Wang, Xin and Alam, Samiul and Wan, Zhongwei and Shen, Hui and Zhang, Mi},
  title = {{SVD-LLM} {V2}: Optimizing Singular Value Truncation for Large Language Model Compression},
  booktitle = {Proceedings of the 2025 Conference of the Nations of the Americas Chapter of the Association for Computational Linguistics: Human Language Technologies (Volume 1: Long Papers)},
  year = {2025},
  pages = {4287--4296},
  publisher = {Association for Computational Linguistics},
  doi = {10.18653/v1/2025.naacl-long.217},
  url = {https://aclanthology.org/2025.naacl-long.217/}
}

@article{safari2026svdsurgeon,
  author = {Safari, Mahmoud and Hutter, Frank},
  title = {{SVD-Surgeon}: Optimal Singular-Value Surgery for Large Language Model Compression},
  journal = {arXiv preprint arXiv:2606.23568},
  year = {2026},
  doi = {10.48550/arXiv.2606.23568},
  url = {https://arxiv.org/abs/2606.23568}
}

@article{jiang2023mistral,
  author = {Jiang, Albert Q. and Sablayrolles, Alexandre and Mensch, Arthur and Bamford, Chris and Chaplot, Devendra Singh and de las Casas, Diego and Bressand, Florian and Lengyel, Gianna and Lample, Guillaume and Saulnier, Lucile and Renard Lavaud, L{\'e}lio and Lachaux, Marie-Anne and Stock, Pierre and Le Scao, Teven and Lavril, Thibaut and Wang, Thomas and Lacroix, Timoth{\'e}e and El Sayed, William},
  title = {Mistral {7B}},
  journal = {arXiv preprint arXiv:2310.06825},
  year = {2023},
  doi = {10.48550/arXiv.2310.06825},
  url = {https://arxiv.org/abs/2310.06825}
}

@article{cobbe2021gsm8k,
  author = {Cobbe, Karl and Kosaraju, Vineet and Bavarian, Mohammad and Chen, Mark and Jun, Heewoo and Kaiser, Lukasz and Plappert, Matthias and Tworek, Jerry and Hilton, Jacob and Nakano, Reiichiro and Hesse, Christopher and Schulman, John},
  title = {Training Verifiers to Solve Math Word Problems},
  journal = {arXiv preprint arXiv:2110.14168},
  year = {2021},
  doi = {10.48550/arXiv.2110.14168},
  url = {https://arxiv.org/abs/2110.14168}
}

@article{chen2021humaneval,
  author = {Chen, Mark and Tworek, Jerry and Jun, Heewoo and Yuan, Qiming and {Ponde de Oliveira Pinto}, Henrique and Kaplan, Jared and Edwards, Harri and Burda, Yuri and Joseph, Nicholas and Brockman, Greg and Ray, Alex and Puri, Raul and Krueger, Gretchen and Petrov, Michael and Khlaaf, Heidy and Sastry, Girish and Mishkin, Pamela and Chan, Brooke and Gray, Scott and Ryder, Nick and Pavlov, Mikhail and Power, Alethea and Kaiser, Lukasz and Bavarian, Mohammad and Winter, Clemens and Tillet, Philippe and {Petroski Such}, Felipe and Cummings, Dave and Plappert, Matthias and Chantzis, Fotios and Barnes, Elizabeth and Herbert-Voss, Ariel and Guss, William Hebgen and Nichol, Alex and Paino, Alex and Tezak, Nikolas and Tang, Jie and Babuschkin, Igor and Balaji, Suchir and Jain, Shantanu and Saunders, William and Hesse, Christopher and Carr, Andrew N. and Leike, Jan and Achiam, Josh and Misra, Vedant and Morikawa, Evan and Radford, Alec and Knight, Matthew and Brundage, Miles and Murati, Mira and Mayer, Katie and Welinder, Peter and McGrew, Bob and Amodei, Dario and McCandlish, Sam and Sutskever, Ilya and Zaremba, Wojciech},
  title = {Evaluating Large Language Models Trained on Code},
  journal = {arXiv preprint arXiv:2107.03374},
  year = {2021},
  doi = {10.48550/arXiv.2107.03374},
  url = {https://arxiv.org/abs/2107.03374}
}

@article{clark2018arc,
  title = {Think you have Solved Question Answering? Try {ARC}, the {AI2} Reasoning Challenge},
  author = {Clark, Peter and Cowhey, Isaac and Etzioni, Oren and Khot, Tushar and Sabharwal, Ashish and Schoenick, Carissa and Tafjord, Oyvind},
  year = {2018},
  journal = {arXiv preprint arXiv:1803.05457},
  doi = {10.48550/arXiv.1803.05457},
  url = {https://arxiv.org/abs/1803.05457}
}

@inproceedings{liu2023evalplus,
  title = {Is Your Code Generated by Chat{GPT} Really Correct? Rigorous Evaluation of Large Language Models for Code Generation},
  author = {Liu, Jiawei and Xia, Chunqiu Steven and Wang, Yuyao and Zhang, Lingming},
  booktitle = {Advances in Neural Information Processing Systems},
  volume = {36},
  year = {2023},
  url = {https://openreview.net/forum?id=1qvx610Cu7}
}

@inproceedings{lecun1989obd,
  author = {LeCun, Yann and Denker, John S. and Solla, Sara A.},
  title = {Optimal Brain Damage},
  booktitle = {Advances in Neural Information Processing Systems},
  volume = {2},
  pages = {598--605},
  year = {1989},
  url = {https://papers.nips.cc/paper_files/paper/1989/hash/6c9882bbac1c7093bd25041881277658-Abstract.html}
}

@inproceedings{hassibi1992obs,
  author = {Hassibi, Babak and Stork, David G.},
  title = {Second Order Derivatives for Network Pruning: Optimal Brain Surgeon},
  booktitle = {Advances in Neural Information Processing Systems},
  volume = {5},
  pages = {164--171},
  year = {1992},
  url = {https://papers.nips.cc/paper_files/paper/1992/hash/303ed4c69846ab36c2904d3ba8573050-Abstract.html}
}

@inproceedings{frantar2023sparsegpt,
  author = {Frantar, Elias and Alistarh, Dan},
  title = {{SparseGPT}: Massive Language Models Can Be Accurately Pruned in One-Shot},
  booktitle = {Proceedings of the 40th International Conference on Machine Learning},
  pages = {10323--10337},
  year = {2023},
  volume = {202},
  series = {Proceedings of Machine Learning Research},
  publisher = {PMLR},
  url = {https://proceedings.mlr.press/v202/frantar23a.html}
}

@inproceedings{sun2024wanda,
  author = {Sun, Mingjie and Liu, Zhuang and Bair, Anna and Kolter, J. Zico},
  title = {A Simple and Effective Pruning Approach for Large Language Models},
  booktitle = {International Conference on Learning Representations},
  year = {2024},
  url = {https://proceedings.iclr.cc/paper_files/paper/2024/hash/14c856c7a41297804de4c4890e846b25-Abstract-Conference.html}
}

@inproceedings{vanderouderaa2024llmsurgeon,
  author = {van der Ouderaa, Tycho F. A. and Nagel, Markus and van Baalen, Mart and Asano, Yuki M. and Blankevoort, Tijmen},
  title = {The {LLM} Surgeon},
  booktitle = {International Conference on Learning Representations},
  year = {2024},
  url = {https://proceedings.iclr.cc/paper_files/paper/2024/file/38a1671ab0747b6ffe4d1c6ef117a3a9-Paper-Conference.pdf}
}

@inproceedings{martens2015kfac,
  author = {Martens, James and Grosse, Roger},
  title = {Optimizing Neural Networks with Kronecker-Factored Approximate Curvature},
  booktitle = {Proceedings of the 32nd International Conference on Machine Learning},
  pages = {2408--2417},
  year = {2015},
  volume = {37},
  series = {Proceedings of Machine Learning Research},
  publisher = {PMLR},
  url = {https://proceedings.mlr.press/v37/martens15.html}
}

@misc{mistral2024configuration,
  author = {{Mistral AI}},
  title = {{Mistral-7B-Instruct-v0.3}: Official Configuration},
  year = {2024},
  howpublished = {Hugging Face model repository},
  url = {https://huggingface.co/mistralai/Mistral-7B-Instruct-v0.3/blob/main/config.json},
  note = {Accessed October 7, 2026; metadata inspection only}
}

@misc{mistral2024weightindex,
  author = {{Mistral AI}},
  title = {{Mistral-7B-Instruct-v0.3}: Official Safetensors Weight Index},
  year = {2024},
  howpublished = {Hugging Face model repository},
  url = {https://huggingface.co/mistralai/Mistral-7B-Instruct-v0.3/blob/main/model.safetensors.index.json},
  note = {Accessed October 7, 2026; parameter inventory only}
}

@misc{transformers2026mistral,
  author = {{Hugging Face}},
  title = {{Transformers}: Official {Mistral} Model Implementation},
  year = {2026},
  howpublished = {Source code},
  url = {https://github.com/huggingface/transformers/blob/main/src/transformers/models/mistral/modeling_mistral.py},
  note = {Accessed October 7, 2026}
}

@article{austin2021mbpp,
  author = {Austin, Jacob and Odena, Augustus and Nye, Maxwell and Bosma, Maarten and Michalewski, Henryk and Dohan, David and Jiang, Ellen and Cai, Carrie and Terry, Michael and Le, Quoc and Sutton, Charles},
  title = {Program Synthesis with Large Language Models},
  journal = {arXiv preprint arXiv:2108.07732},
  year = {2021},
  doi = {10.48550/arXiv.2108.07732},
  url = {https://arxiv.org/abs/2108.07732}
}

@article{dubey2024llama3,
  author = {Dubey, Abhimanyu and Jauhri, Abhinav and Pandey, Abhinav and Kadian, Abhishek and Al-Dahle, Ahmad and Letman, Aiesha and Mathur, Akhil and Schelten, Alan and Yang, Amy and Fan, Angela and Goyal, Anirudh and Hartshorn, Anthony and Yang, Aobo and Mitra, Archi and Sravankumar, Archie and Korenev, Artem and Hinsvark, Arthur and Rao, Arun and Zhang, Aston and Rodriguez, Aurelien and Gregerson, Austen and Spataru, Ava and Roziere, Baptiste and Biron, Bethany and Tang, Binh and Chern, Bobbie and Caucheteux, Charlotte and Nayak, Chaya and Bi, Chloe and Marra, Chris and McConnell, Chris and Keller, Christian and Touret, Christophe and Wu, Chunyang and Wong, Corinne and Ferrer, Cristian Canton and Nikolaidis, Cyrus and Allonsius, Damien and Song, Daniel and Pintz, Danielle and Livshits, Danny and Esiobu, David and Choudhary, Dhruv and Mahajan, Dhruv and Garcia-Olano, Diego and Perino, Diego and Hupkes, Dieuwke and Lakomkin, Egor and AlBadawy, Ehab and Lobanova, Elina and Dinan, Emily and Smith, Eric Michael and Radenovic, Filip and Zhang, Frank and Synnaeve, Gabriel and Lee, Gabrielle and Anderson, Georgia Lewis and Nail, Graeme and Mialon, Gregoire and Pang, Guan and Cucurell, Guillem and Nguyen, Hailey and Korevaar, Hannah and Xu, Hu and Touvron, Hugo and Zarov, Iliyan and Ibarra, Imanol Arrieta and Kloumann, Isabel and Misra, Ishan and Evtimov, Ivan and Copet, Jade and Lee, Jaewon and Geffert, Jan and Vranes, Jana and Park, Jason and Mahadeokar, Jay and Shah, Jeet and van der Linde, Jelmer and Billock, Jennifer and Hong, Jenny and Lee, Jenya and Fu, Jeremy and Chi, Jianfeng and Huang, Jianyu and Liu, Jiawen and Wang, Jie and Yu, Jiecao and Bitton, Joanna and Spisak, Joe and Park, Jongsoo and Rocca, Joseph and Johnstun, Joshua and Saxe, Joshua and Jia, Junteng and Alwala, Kalyan Vasuden and Upasani, Kartikeya and Plawiak, Kate and Li, Ke and Heafield, Kenneth and Stone, Kevin and El-Arini, Khalid and Iyer, Krithika and Malik, Kshitiz and Chiu, Kuenley and Bhalla, Kunal and Rantala-Yeary, Lauren and van der Maaten, Laurens and Chen, Lawrence and Tan, Liang and Jenkins, Liz and Martin, Louis and Madaan, Lovish and Malo, Lubo and Blecher, Lukas and Landzaat, Lukas and de Oliveira, Luke and Muzzi, Madeline and Pasupuleti, Mahesh and Singh, Mannat and Paluri, Manohar and Kardas, Marcin and Oldham, Mathew and Rita, Mathieu and Pavlova, Maya and Kambadur, Melanie and Lewis, Mike and Si, Min and Singh, Mitesh Kumar and Hassan, Mona and Goyal, Naman and Torabi, Narjes and Bashlykov, Nikolay and Bogoychev, Nikolay and Chatterji, Niladri and Duchenne, Olivier and {\c{C}}elebi, Onur and Alrassy, Patrick and Zhang, Pengchuan and Li, Pengwei and Vasic, Petar and Weng, Peter and Bhargava, Prajjwal and Dubal, Pratik and Krishnan, Praveen and Koura, Punit Singh and Xu, Puxin and He, Qing and Dong, Qingxiao and Srinivasan, Ragavan and Ganapathy, Raj and Calderer, Ramon and Cabral, Ricardo Silveira and Stojnic, Robert and Raileanu, Roberta and Girdhar, Rohit and Patel, Rohit and Sauvestre, Romain and Polidoro, Ronnie and Sumbaly, Roshan and Taylor, Ross and Silva, Ruan and Hou, Rui and Wang, Rui and Hosseini, Saghar and Chennabasappa, Sahana and Singh, Sanjay and Bell, Sean and Kim, Seohyun Sonia and Edunov, Sergey and Nie, Shaoliang and Narang, Sharan and Raparthy, Sharath and Shen, Sheng and Wan, Shengye and Bhosale, Shruti and Zhang, Shun and Vandenhende, Simon and Batra, Soumya and Whitman, Spencer and Sootla, Sten and Collot, Stephane and Gururangan, Suchin and Borodinsky, Sydney and Herman, Tamar and Fowler, Tara and Sheasha, Tarek and Georgiou, Thomas and Scialom, Thomas and Speckbacher, Tobias and Mihaylov, Todor and Xiao, Tong and Karn, Ujjwal and Goswami, Vedanuj and Gupta, Vibhor and Ramanathan, Vignesh and Kerkez, Viktor and Gonguet, Vincent and Do, Virginie and Vogeti, Vish and Petrovic, Vladan and Chu, Weiwei and Xiong, Wenhan and Fu, Wenyin and Meers, Whitney and Martinet, Xavier and Wang, Xiaodong and Tan, Xiaoqing Ellen and Xie, Xinfeng and Jia, Xuchao and Wang, Xuewei and Goldschlag, Yaelle and Gaur, Yashesh and Babaei, Yasmine and Wen, Yi and Song, Yiwen and Zhang, Yuchen and Li, Yue and Mao, Yuning and Coudert, Zacharie Delpierre and Yan, Zheng and Chen, Zhengxing and Papakipos, Zoe and Singh, Aaditya and Grattafiori, Aaron and Jain, Abha and Kelsey, Adam and Shajnfeld, Adam and Gangidi, Adithya and Victoria, Adolfo and Goldstand, Ahuva and Menon, Ajay and Sharma, Ajay and Boesenberg, Alex and Vaughan, Alex and Baevski, Alexei and Feinstein, Allie and Kallet, Amanda and Sangani, Amit and Yunus, Anam and Lupu, Andrei and Alvarado, Andres and Caples, Andrew and Gu, Andrew and Ho, Andrew and Poulton, Andrew and Ryan, Andrew and Ramchandani, Ankit and Franco, Annie and Saraf, Aparajita and Chowdhury, Arkabandhu and Gabriel, Ashley and Bharambe, Ashwin and Eisenman, Assaf and Yazdan, Azadeh and James, Beau and Maurer, Ben and Leonhardi, Benjamin and Huang, Bernie and Loyd, Beth and De Paola, Beto and Paranjape, Bhargavi and Liu, Bing and Wu, Bo and Ni, Boyu and Hancock, Braden and Wasti, Bram and Spence, Brandon and Stojkovic, Brani and Gamido, Brian and Montalvo, Britt and Parker, Carl and Burton, Carly and Mejia, Catalina and Wang, Changhan and Kim, Changkyu and Zhou, Chao and Hu, Chester and Chu, Ching-Hsiang and Cai, Chris and Tindal, Chris and Feichtenhofer, Christoph and Civin, Damon and Beaty, Dana and Kreymer, Daniel and Li, Daniel and Wyatt, Danny and Adkins, David and Xu, David and Testuggine, Davide and David, Delia and Parikh, Devi and Liskovich, Diana and Foss, Didem and Wang, Dingkang and Le, Duc and Holland, Dustin and Dowling, Edward and Jamil, Eissa and Montgomery, Elaine and Presani, Eleonora and Hahn, Emily and Wood, Emily and Brinkman, Erik and Arcaute, Esteban and Dunbar, Evan and Smothers, Evan and Sun, Fei and Kreuk, Felix and Tian, Feng and Ozgenel, Firat and Caggioni, Francesco and Guzm{\'a}n, Francisco and Kanayet, Frank and Seide, Frank and Florez, Gabriela Medina and Schwarz, Gabriella and Badeer, Gada and Swee, Georgia and Halpern, Gil and Thattai, Govind and Herman, Grant and Sizov, Grigory and Zhang, Guangyi and Lakshminarayanan, Guna and Shojanazeri, Hamid and Zou, Han and Wang, Hannah and Zha, Hanwen and Habeeb, Haroun and Rudolph, Harrison and Suk, Helen and Aspegren, Henry and Goldman, Hunter and Molybog, Igor and Tufanov, Igor and Veliche, Irina-Elena and Gat, Itai and Weissman, Jake and Geboski, James and Kohli, James and Asher, Japhet and Gaya, Jean-Baptiste and Marcus, Jeff and Tang, Jeff and Chan, Jennifer and Zhen, Jenny and Reizenstein, Jeremy and Teboul, Jeremy and Zhong, Jessica and Jin, Jian and Yang, Jingyi and Cummings, Joe and Carvill, Jon and Shepard, Jon and McPhie, Jonathan and Torres, Jonathan and Ginsburg, Josh and Wang, Junjie and Wu, Kai and U, Kam Hou and Saxena, Karan and Prasad, Karthik and Khandelwal, Kartikay and Zand, Katayoun and Matosich, Kathy and Veeraraghavan, Kaushik and Michelena, Kelly and Li, Keqian and Huang, Kun and Chawla, Kunal and Lakhotia, Kushal and Huang, Kyle and Chen, Lailin and Garg, Lakshya and A, Lavender and Silva, Leandro and Bell, Lee and Zhang, Lei and Guo, Liangpeng and Yu, Licheng and Moshkovich, Liron and Wehrstedt, Luca and Khabsa, Madian and Avalani, Manav and Bhatt, Manish and Tsimpoukelli, Maria and Mankus, Martynas and Hasson, Matan and Lennie, Matthew and Reso, Matthias and Groshev, Maxim and Naumov, Maxim and Lathi, Maya and Keneally, Meghan and Seltzer, Michael L. and Valko, Michal and Restrepo, Michelle and Patel, Mihir and Vyatskov, Mik and Samvelyan, Mikayel and Clark, Mike and Macey, Mike and Wang, Mike and Hermoso, Miquel Jubert and Metanat, Mo and Rastegari, Mohammad and Bansal, Munish and Santhanam, Nandhini and Parks, Natascha and White, Natasha and Bawa, Navyata and Singhal, Nayan and Egebo, Nick and Usunier, Nicolas and Laptev, Nikolay Pavlovich and Dong, Ning and Zhang, Ning and Cheng, Norman and Chernoguz, Oleg and Hart, Olivia and Salpekar, Omkar and Kalinli, Ozlem and Kent, Parkin and Parekh, Parth and Saab, Paul and Balaji, Pavan and Rittner, Pedro and Bontrager, Philip and Roux, Pierre and Dollar, Piotr and Zvyagina, Polina and Ratanchandani, Prashant and Yuvraj, Pritish and Liang, Qian and Alao, Rachad and Rodriguez, Rachel and Ayub, Rafi and Murthy, Raghotham and Nayani, Raghu and Mitra, Rahul and Li, Raymond and Hogan, Rebekkah and Battey, Robin and Wang, Rocky and Maheswari, Rohan and Howes, Russ and Rinott, Ruty and Bondu, Sai Jayesh and Datta, Samyak and Chugh, Sara and Hunt, Sara and Dhillon, Sargun and Sidorov, Sasha and Pan, Satadru and Verma, Saurabh and Yamamoto, Seiji and Ramaswamy, Sharadh and Lindsay, Shaun and Lindsay, Shaun and Feng, Sheng and Lin, Shenghao and Zha, Shengxin Cindy and Shankar, Shiva and Zhang, Shuqiang and Zhang, Shuqiang and Wang, Sinong and Agarwal, Sneha and Sajuyigbe, Soji and Chintala, Soumith and Max, Stephanie and Chen, Stephen and Kehoe, Steve and Satterfield, Steve and Govindaprasad, Sudarshan and Gupta, Sumit and Cho, Sungmin and Virk, Sunny and Subramanian, Suraj and Choudhury, Sy and Goldman, Sydney and Remez, Tal and Glaser, Tamar and Best, Tamara and Kohler, Thilo and Robinson, Thomas and Li, Tianhe and Zhang, Tianjun and Matthews, Tim and Chou, Timothy and Shaked, Tzook and Vontimitta, Varun and Ajayi, Victoria and Montanez, Victoria and Mohan, Vijai and Kumar, Vinay Satish and Mangla, Vishal and Ionescu, Vlad and Poenaru, Vlad and Mihailescu, Vlad Tiberiu and Ivanov, Vladimir and Li, Wei and Wang, Wenchen and Jiang, Wenwen and Bouaziz, Wes and Constable, Will and Tang, Xiaocheng and Wang, Xiaofang and Wu, Xiaojian and Wang, Xiaolan and Xia, Xide and Wu, Xilun and Gao, Xinbo and Chen, Yanjun and Hu, Ye and Jia, Ye and Qi, Ye and Li, Yenda and Zhang, Yilin and Zhang, Ying and Adi, Yossi and Nam, Youngjin and Wang, Yu and Hao, Yuchen and Qian, Yundi and He, Yuzi and Rait, Zach and DeVito, Zachary and Rosnbrick, Zef and Wen, Zhaoduo and Yang, Zhenyu and Zhao, Zhiwei},
  title = {The {Llama} 3 Herd of Models},
  journal = {arXiv preprint arXiv:2407.21783},
  year = {2024},
  doi = {10.48550/arXiv.2407.21783},
  url = {https://arxiv.org/abs/2407.21783v1}
}

@misc{mistral2024modelcard,
  author = {{Mistral AI}},
  title = {{Mistral-7B-Instruct-v0.3}},
  year = {2024},
  howpublished = {Official model card, Hugging Face},
  url = {https://huggingface.co/mistralai/Mistral-7B-Instruct-v0.3},
  note = {Accessed October 11, 2026}
}

@misc{meta2024llama31instruct,
  author = {{Meta}},
  title = {{Llama-3.1-8B-Instruct}},
  year = {2024},
  howpublished = {Official model card, Hugging Face},
  url = {https://huggingface.co/meta-llama/Llama-3.1-8B-Instruct},
  note = {Accessed October 11, 2026}
}

@inproceedings{yao2024llmunlearning,
  author = {Yao, Yuanshun and Xu, Xiaojun and Liu, Yang},
  title = {Large Language Model Unlearning},
  booktitle = {Advances in Neural Information Processing Systems},
  volume = {37},
  year = {2024},
  url = {https://openreview.net/forum?id=8Dy42ThoNe}
}

@inproceedings{zhang2024npo,
  author = {Zhang, Ruiqi and Lin, Licong and Bai, Yu and Mei, Song},
  title = {Negative Preference Optimization: From Catastrophic Collapse to Effective Unlearning},
  booktitle = {First Conference on Language Modeling},
  year = {2024},
  url = {https://openreview.net/forum?id=MXLBXjQkmb}
}

@inproceedings{maini2024tofu,
  author = {Maini, Pratyush and Feng, Zhili and Schwarzschild, Avi and Lipton, Zachary C. and Kolter, J. Zico},
  title = {{TOFU}: A Task of Fictitious Unlearning for {LLMs}},
  booktitle = {First Conference on Language Modeling},
  year = {2024},
  url = {https://openreview.net/forum?id=B41hNBoWLo}
}

@inproceedings{shi2025muse,
  author = {Shi, Weijia and Lee, Jaechan and Huang, Yangsibo and Malladi, Sadhika and Zhao, Jieyu and Holtzman, Ari and Liu, Daogao and Zettlemoyer, Luke and Smith, Noah A. and Zhang, Chiyuan},
  title = {{MUSE}: Machine Unlearning Six-Way Evaluation for Language Models},
  booktitle = {International Conference on Learning Representations},
  year = {2025},
  url = {https://openreview.net/forum?id=TArmA033BU}
}

@article{eldan2023harry,
  author = {Eldan, Ronen and Russinovich, Mark},
  title = {Who's {Harry Potter}? Approximate Unlearning in {LLMs}},
  journal = {arXiv preprint arXiv:2310.02238},
  year = {2023},
  doi = {10.48550/arXiv.2310.02238},
  url = {https://arxiv.org/abs/2310.02238}
}

@inproceedings{jang2023knowledge,
  author = {Jang, Joel and Yoon, Dongkeun and Yang, Sohee and Cha, Sungmin and Lee, Moontae and Logeswaran, Lajanugen and Seo, Minjoon},
  title = {Knowledge Unlearning for Mitigating Privacy Risks in Language Models},
  booktitle = {Proceedings of the 61st Annual Meeting of the Association for Computational Linguistics (Volume 1: Long Papers)},
  year = {2023},
  pages = {14389--14408},
  publisher = {Association for Computational Linguistics},
  doi = {10.18653/v1/2023.acl-long.805},
  url = {https://aclanthology.org/2023.acl-long.805/}
}

@article{liu2025rethinking,
  author = {Liu, Sijia and Yao, Yuanshun and Jia, Jinghan and Casper, Stephen and Baracaldo, Nathalie and Hase, Peter and Yao, Yuguang and Liu, Chris Yuhao and Xu, Xiaojun and Li, Hang and Varshney, Kush R. and Bansal, Mohit and Koyejo, Sanmi and Liu, Yang},
  title = {Rethinking Machine Unlearning for Large Language Models},
  journal = {Nature Machine Intelligence},
  volume = {7},
  number = {2},
  pages = {181--194},
  year = {2025},
  doi = {10.1038/s42256-025-00985-0}
}

@inproceedings{arditi2024refusal,
  author = {Arditi, Andy and Obeso, Oscar and Syed, Aaquib and Paleka, Daniel and Panickssery, Nina and Gurnee, Wes and Nanda, Neel},
  title = {Refusal in Language Models Is Mediated by a Single Direction},
  booktitle = {Advances in Neural Information Processing Systems},
  volume = {37},
  year = {2024},
  url = {https://openreview.net/forum?id=pH3XAQME6c}
}

@inproceedings{wei2024brittleness,
  author = {Wei, Boyi and Huang, Kaixuan and Huang, Yangsibo and Xie, Tinghao and Qi, Xiangyu and Xia, Mengzhou and Mittal, Prateek and Wang, Mengdi and Henderson, Peter},
  title = {Assessing the Brittleness of Safety Alignment via Pruning and Low-Rank Modifications},
  booktitle = {Proceedings of the 41st International Conference on Machine Learning},
  year = {2024},
  volume = {235},
  series = {Proceedings of Machine Learning Research},
  pages = {52588--52610},
  publisher = {PMLR},
  url = {https://proceedings.mlr.press/v235/wei24f.html}
}

@inproceedings{hu2025jogging,
  author = {Hu, Shengyuan and Fu, Yiwei and Wu, Zhiwei Steven and Smith, Virginia},
  title = {Unlearning or Obfuscating? Jogging the Memory of Unlearned {LLMs} via Benign Relearning},
  booktitle = {International Conference on Learning Representations},
  year = {2025},
  url = {https://openreview.net/forum?id=fMNRYBvcQN}
}

@article{lucki2025adversarial,
  author = {{\L}ucki, Jakub and Wei, Boyi and Huang, Yangsibo and Henderson, Peter and Tram{\`e}r, Florian and Rando, Javier},
  title = {An Adversarial Perspective on Machine Unlearning for {AI} Safety},
  journal = {Transactions on Machine Learning Research},
  year = {2025},
  url = {https://openreview.net/forum?id=J5IRyTKZ9s}
}

@inproceedings{hsu2022fwsvd,
  author = {Hsu, Yen-Chang and Hua, Ting and Chang, Sungen and Lou, Qian and Shen, Yilin and Jin, Hongxia},
  title = {Language Model Compression with Weighted Low-Rank Factorization},
  booktitle = {International Conference on Learning Representations},
  year = {2022},
  url = {https://arxiv.org/abs/2207.00112}
}

@article{yuan2023asvd,
  author = {Yuan, Zhihang and Shang, Yuzhang and Song, Yue and Yang, Dawei and Wu, Qiang and Yan, Yan and Sun, Guangyu},
  title = {{ASVD}: Activation-aware Singular Value Decomposition for Compressing Large Language Models},
  journal = {arXiv preprint arXiv:2312.05821},
  year = {2023},
  doi = {10.48550/arXiv.2312.05821},
  url = {https://arxiv.org/abs/2312.05821}
}

@article{pochinkov2024dissecting,
  author = {Pochinkov, Nicholas and Schoots, Nandi},
  title = {Dissecting Language Models: Machine Unlearning via Selective Pruning},
  journal = {arXiv preprint arXiv:2403.01267},
  year = {2024},
  doi = {10.48550/arXiv.2403.01267},
  url = {https://arxiv.org/abs/2403.01267}
}

@inproceedings{ortizjimenez2023tangent,
  author = {Ortiz-Jimenez, Guillermo and Favero, Alessandro and Frossard, Pascal},
  title = {Task Arithmetic in the Tangent Space: Improved Editing of Pre-Trained Models},
  booktitle = {Advances in Neural Information Processing Systems},
  volume = {36},
  year = {2023},
  url = {https://openreview.net/forum?id=0A9f2jZDGW}
}

@inproceedings{koh2017influence,
  author = {Koh, Pang Wei and Liang, Percy},
  title = {Understanding Black-box Predictions via Influence Functions},
  booktitle = {Proceedings of the 34th International Conference on Machine Learning},
  year = {2017},
  volume = {70},
  series = {Proceedings of Machine Learning Research},
  pages = {1885--1894},
  publisher = {PMLR},
  url = {https://proceedings.mlr.press/v70/koh17a.html}
}

@inproceedings{basu2021fragile,
  author = {Basu, Samyadeep and Pope, Philip and Feizi, Soheil},
  title = {Influence Functions in Deep Learning Are Fragile},
  booktitle = {International Conference on Learning Representations},
  year = {2021},
  url = {https://arxiv.org/abs/2006.14651}
}

@article{lei2018distribution,
  author = {Lei, Jing and G'Sell, Max and Rinaldo, Alessandro and Tibshirani, Ryan J. and Wasserman, Larry},
  title = {Distribution-Free Predictive Inference for Regression},
  journal = {Journal of the American Statistical Association},
  volume = {113},
  number = {523},
  pages = {1094--1111},
  year = {2018},
  doi = {10.1080/01621459.2017.1307116}
}
\clearpage
\appendix
\raggedbottom
\makeatletter
\setlength{\@fptop}{0pt}
\setlength{\@fpsep}{12pt}
\setlength{\@fpbot}{0pt plus 1fil}
\setlength{\@dblfptop}{0pt}
\setlength{\@dblfpsep}{12pt}
\setlength{\@dblfpbot}{0pt plus 1fil}
\makeatother
\renewcommand{\topfraction}{0.9}
\renewcommand{\textfraction}{0.1}
\renewcommand{\floatpagefraction}{0.85}
\renewcommand{\dbltopfraction}{0.9}
\renewcommand{\dblfloatpagefraction}{0.9}
\setcounter{topnumber}{3}
\setcounter{dbltopnumber}{3}
\setcounter{totalnumber}{4}
\setlength{\textfloatsep}{12pt plus 2pt minus 2pt}
\setlength{\floatsep}{12pt plus 2pt minus 2pt}
\setlength{\dbltextfloatsep}{12pt plus 2pt minus 2pt}
\setlength{\dblfloatsep}{12pt plus 2pt minus 2pt}

\section{Notation}\label{app:notation}

Table~\ref{tab:notation} summarizes notation used in the main text and appendix. Superscripts $\mathrm{search}$, $\mathrm{val}$, and $\mathrm{test}$ identify item splits. Hats denote empirical or numerical estimates. Indices are local to each definition: $j$ indexes a singular component in $(m,j)$ and a candidate in the screening analysis.

Support labels such as A, B, and C identify selected masks in the reported results.

\begin{table*}[!tp]
\centering
\small
\setlength{\tabcolsep}{5pt}
\renewcommand{\arraystretch}{1.08}
\begin{tabular}{@{}>{\raggedright\arraybackslash}p{0.23\textwidth}>{\raggedright\arraybackslash}p{\dimexpr0.77\textwidth-2\tabcolsep\relax}@{}}
\toprule
Symbol & Definition\\
\midrule
\multicolumn{2}{@{}l}{\textit{Model and spectral intervention}}\\
\addlinespace[0.2em]
$f_\theta,\theta$ & Language model and unedited parameters.\\
$\calM,W_m$ & Editable matrices and matrix $m$, of shape $d_m^{\rm out}\times d_m^{\rm in}$.\\
$\sigma_{m,j},u_{m,j},v_{m,j}$ & Singular value and left/right singular vectors of $W_m$; $r_m=\min(d_m^{\rm out},d_m^{\rm in})$ is the thin-SVD size, which can exceed numerical rank.\\
$\mathcal J_m,\calI,N$ & Searchable indices in matrix $m$, pooled coordinate set, and size $N=|\calI|$.\\
$i,A_i$ & Coordinate $i=(m,j)$ and atom $A_i=\sigma_{m,j}u_{m,j}v_{m,j}^{\top}$ in block $m$.\\
$S,d_S,\alpha$ & Selected coordinates, direction $d_S=-\sum_{i\in S}A_i$, and amplitude $\alpha\in[0,1]$; the edit is $\theta+\alpha d_S$, with full deletion at $\alpha=1$.\\
$R(S),B$ & Coordinate count $R(S)=|S|$ and budget; this count differs from matrix-update rank.\\
$e_i,E(S),E_{\max}$ & Coordinate energy $e_i=\sigma_{m,j}^2$, deletion energy $E(S)=\norm{d_S}_F^2$, and cap.\\
$E_{\rm band},\bar E$ & Searchable energy $E_{\rm band}=\sum_{i\in\calI}e_i$ and normalized energy $\bar E=E/E_{\rm band}$.\\
$\ip{X}{Y},\norm{X}_F$ & Frobenius inner product and norm, summed over editable blocks for parameter inner products and squared norms.\\
\addlinespace[0.35em]
\multicolumn{2}{@{}l}{\textit{Tasks, losses, and local predictions}}\\
\addlinespace[0.2em]
$\calC,c,t,P,p$ & Capability set, task index, target task, preserved-task set, and task $p\in P$.\\
$\calD_c,\ell_c,L_c$ & Task distribution, item loss, and expected loss; empirical losses average item losses.\\
$\Delta_c(S,\alpha)$ & Loss change from the unedited model; positive values indicate degradation.\\
$\operatorname{Acc}_c,D_c(S,\alpha)$ & Higher-is-better task score and its drop, separate from the differentiable loss and its change.\\
$\tau_t,\epsilon_p$ & Target-loss increase floor and preservation tolerance: $|\Delta_p|\leq\epsilon_p$ for absolute preservation, $\Delta_p\leq\epsilon_p$ for harm-only preservation.\\
$G_{c,m},\gamma_{c,i}$ & Unedited matrix gradient and signed coordinate-deletion derivative in the frozen singular basis (Equation~\ref{eq:gamma}).\\
$P_c(S,\alpha),P_{c,j}$ & First-order prediction (Equation~\ref{eq:measured-path}); superscripts identify derivative splits. $P_c(S)$ denotes full deletion, and $P_{c,j}$ is candidate $j$'s search-split full-deletion prediction.\\
$\mathcal R_c,\widehat{\mathcal R}_c$ & Exact Taylor remainder and measured same-split prediction residual.\\
$r_c^{\rm val}(S)$ & Validation loss change minus search prediction; includes derivative transfer and the validation Taylor remainder (Equation~\ref{eq:validation-error}).\\
\addlinespace[0.35em]
\multicolumn{2}{@{}l}{\textit{Curvature, calibration, and screening}}\\
\addlinespace[0.2em]
$g_c(\alpha),h_{c,S}(u)$ & Path loss and second directional derivative at $u$ (Equation~\ref{eq:taylor}).\\
$H_c,\bar H$ & Conditional bound coefficient and normalization $\bar H=HE_{\rm band}$; requires $|h_{c,S}(u)|\leq H_cE(S)$ along the path.\\
$s,\bar H_{c,s},\kappa$ & Energy stratum, fitted empirical-quantile coefficient, and calibration multiplier; the coefficient is not a certified curvature bound.\\
$Q_q$ & Empirical quantile at level $q$.\\
$A_{c,j}$ & Nonnegative error radius for task $c$ and candidate $j$; calibration uses $A_c(S;\kappa)$ (Equation~\ref{eq:calibration}).\\
$I_{c,j},\mathcal U_j$ & Interval $[P_{c,j}-A_{c,j},P_{c,j}+A_{c,j}]$ and product of task intervals.\\
$Q,j^*$ & Verification cap and first feasible candidate in the fixed order; $Q$ differs from $Q_q$.\\
\addlinespace[0.35em]
\multicolumn{2}{@{}l}{\textit{Appendix diagnostics and local conventions}}\\
\addlinespace[0.2em]
$q_c,I_c$ & Full-amplitude base-point quadratic term and numerically integrated test-path curvature; $I_c$ differs from $I_{c,j}$.\\
$K_{c,S},K_c$ & Pathwise third-derivative bound and Hessian Lipschitz constant in operator norm, under the stated assumptions.\\
$T_c,A_c,M_{2,c}$ & Signed, absolute, and squared task-score sums; this $A_c$ differs from $A_c(S;\kappa)$ and $A_i$.\\
$N_{\rm eff,c},N_{\rm eff,E}$ & Effective magnitude counts for task scores and coordinate energies.\\
$a_c,b_c,u_{c,j}$ & Scaling-fit intercept, slope, and residual; the proposal generator uses $a_c$ for task-score normalization and $z_i$ for a Gumbel perturbation.\\
\bottomrule
\end{tabular}
\caption{Notation used in the main text and appendix.}\label{tab:notation}
\end{table*}

\section{Proofs}\label{app:proofs}

\subsection{Spectral norms and the frozen-coordinate derivative}

For each matrix, distinct SVD components have orthogonal left and right singular vectors. Their Frobenius inner product is zero because
\begin{equation}
 \ip{u_jv_j^\top}{u_kv_k^\top}
 =(u_j^\top u_k)(v_j^\top v_k).
\end{equation}
The selected components therefore have squared Frobenius norm $\sum_{j\in S_m}\sigma_{m,j}^2$, operator norm $\max_{j\in S_m}\sigma_{m,j}$, and rank equal to the number of selected nonzero singular values. Empty selections have norm and rank zero. Distinct parameter matrices occupy disjoint blocks, giving Equation~\ref{eq:energy}. Budget calculations exclude duplicate support entries.

For a fixed singular basis, differentiation along the deletion direction gives
\begin{equation}
\begin{aligned}
 g_c'(0)&=\sum_m\ip{G_{c,m}}{-\sum_{j\in S_m}A_{m,j}}\\
         &=\sum_{(m,j)\in S}-\sigma_{m,j}u_{m,j}^{\top}G_{c,m}v_{m,j}.
\end{aligned}
\end{equation}
The analysis computes the basis once from the unedited matrices and holds it fixed during differentiation. Recomputing it along the path defines a different intervention.

\subsection{Taylor identity and finite-amplitude error}

The fundamental theorem of calculus yields
\begin{equation}
 g_c(\alpha)-g_c(0)=\int_0^\alpha g_c'(u)\,du.
\end{equation}
Substituting $g_c'(u)=g_c'(0)+\int_0^u g_c''(v)\,dv$ and changing the order of integration gives
\begin{equation}
 g_c(\alpha)-g_c(0)
 =\alpha g_c'(0)+\int_0^\alpha(\alpha-v)g_c''(v)\,dv.
\end{equation}
Since $g_c''(v)=d_S^\top\nabla^2L_c(\theta+v d_S)d_S$, this proves Equation~\ref{eq:taylor}. A uniform directional bound $|g_c''(v)|\leq H_cE(S)$ gives Equation~\ref{eq:remainder} after integrating $\alpha-v$. A global Hessian operator-norm bound over the candidate paths is sufficient, but substantially stronger than a pointwise Hessian estimate at the unedited model.

If a third derivative along the path is uniformly bounded by $K_{c,S}$, then
\begin{equation}
 \left|\Delta_c(S,\alpha)-\alpha g_c'(0)
       -\frac{\alpha^2}{2}g_c''(0)\right|
 \leq\frac{\alpha^3}{6}K_{c,S}.
\end{equation}
Without a bound on higher derivatives or direct path measurements, the linear residual cannot be equated with the base-point quadratic term. This derivation applies to a differentiable loss; binary accuracy has no corresponding Taylor remainder.

\subsection{Base and path curvature}
\label{app:curvature-proof}

For a frozen direction, $P_c^{\rm search}$ and $P_c^{\rm test}$ use the same answer-only NLL on their respective items. At full amplitude, we define $q_c=h_{c,S}^{\rm test}(0)/2$ and $I_c=\int_0^1(1-u)\widehat h_{c,S}^{\rm test}(u)\,du$. Then
\begin{equation}
\begin{aligned}
 \widehat\Delta_c^{\rm test}={}&P_c^{\rm search}
 +(P_c^{\rm test}-P_c^{\rm search})+q_c\\
 &+(I_c-q_c)+(\widehat{\mathcal R}_c-I_c).
\end{aligned}
\label{eq:decomposition}
\end{equation}
The decomposition separates derivative transfer, base-point curvature, and variation along the path. Its final term collects quadrature error, HVP error, and disagreement between objectives or implementations. Refining quadrature addresses only one of these sources.

For general amplitude, $q_{c,S}(\alpha)=\alpha^2h_{c,S}(0)/2$. The exact omitted path term satisfies
\begin{equation}
\begin{aligned}
 &\mathcal R_c(S,\alpha)-q_{c,S}(\alpha)\\
 &\quad=\int_0^\alpha(\alpha-u)
 [h_{c,S}(u)-h_{c,S}(0)]\,du.
\end{aligned}
\label{eq:path-variation}
\end{equation}
For $\omega_{c,S}(\alpha)=\sup_{u\leq\alpha}|h_{c,S}(u)-h_{c,S}(0)|$,
\begin{equation}
 |\mathcal R_c-q_{c,S}|\leq\tfrac12\alpha^2\omega_{c,S}(\alpha).
 \label{eq:variation-bound}
\end{equation}
A $K_c$-Lipschitz Hessian gives the alternative bound $K_c\alpha^3E(S)^{3/2}/6$. A finite HVP grid certifies neither a path supremum nor a Lipschitz constant. The bounds follow from the integral representation.

For a frozen displacement $d_S$, twice continuous differentiability gives Equation~\ref{eq:taylor}. Since $\int_0^\alpha(\alpha-u)\,du=\alpha^2/2$, subtracting $\alpha^2h_{c,S}(0)/2$ yields Equation~\ref{eq:path-variation}. Its absolute value is at most
\begin{equation}
\begin{aligned}
 &\int_0^\alpha(\alpha-u)
  |h_{c,S}(u)-h_{c,S}(0)|\,du\\
 &\hspace{1em}\leq\tfrac12\alpha^2\omega_{c,S}(\alpha),
\end{aligned}
\end{equation}
which proves Equation~\ref{eq:variation-bound} without a sign assumption.

If the Hessian is $K_c$-Lipschitz in operator norm on this segment, then
\begin{equation}
\begin{aligned}
 |h_{c,S}(u)-h_{c,S}(0)|
 &\leq\norm{d_S}^2 K_cu\norm{d_S}\\
 &=K_cuE(S)^{3/2}.
\end{aligned}
\end{equation}
Integrating $u(\alpha-u)$ gives $\alpha^3/6$ and the stated cubic bound. The result remains conditional: a finite measurement grid does not determine the required Lipschitz constant.

Spectral orthogonality simplifies deletion energy, but it does not diagonalize the task Hessian. A Hessian-vector product (HVP) with the full displacement retains cross terms between atoms and matrices. Summing single-coordinate quadratic curvatures omits those terms. The HVP vectors themselves are linear in the displacement and can be summed exactly.

\subsection{Interval pruning and ordered verification}
\label{app:outer-proof}

The screening procedure removes candidates that violate known rank, energy, or support constraints. For each remaining candidate $j$, we define the task interval $I_{c,j}=[P_{c,j}-A_{c,j},P_{c,j}+A_{c,j}]$, with finite $A_{c,j}\geq0$, and the uncertainty rectangle $\mathcal U_j=\prod_c I_{c,j}$. The target acceptance set is $[\tau_t,\infty)$. A preserved task accepts $[-\epsilon_p,\epsilon_p]$ under absolute stability or $(-\infty,\epsilon_p]$ under harm-only preservation, with $\epsilon_p\geq0$.

The inner test requires each interval to lie entirely inside its acceptance set. Its conditions are $P_{t,j}-A_{t,j}\geq\tau_t$ and $|P_{p,j}|+A_{p,j}\leq\epsilon_p$, or $P_{p,j}+A_{p,j}\leq\epsilon_p$ for harm-only preservation. Passing this test makes every loss vector in the rectangle feasible. Failing it does not imply that the rectangle contains no feasible vector.

The target interval intersects its acceptance half-line precisely when $P_{t,j}+A_{t,j}\geq\tau_t$. A preserved absolute-stability interval intersects its acceptance segment precisely when
\begin{equation}
\begin{aligned}
 P_{p,j}-A_{p,j}&\leq\epsilon_p,\\
 P_{p,j}+A_{p,j}&\geq-\epsilon_p.
\end{aligned}
\end{equation}
These inequalities are equivalent to $|P_{p,j}|-A_{p,j}\leq\epsilon_p$. Under harm-only preservation, intersection requires only $P_{p,j}-A_{p,j}\leq\epsilon_p$. An actually feasible value lies in both the interval and the acceptance set, so it satisfies every outer inequality. Therefore no feasible candidate is pruned under the stated envelopes.

The verification procedure keeps the pool in the fixed direct-check order. Its first feasible candidate $j^*$ survives outer screening. Each earlier survivor is infeasible and fails the same finite verification, so screening stops at $j^*$ after at most $j^*$ checks. If the pool has no feasible candidate, both policies abstain and screening checks at most the full pool. An empty pool requires zero checks.

The verification procedure resolves candidate-ID ties before screening and preserves survivor order. Reranking survivors by a reserved target score invalidates the guarantee that both policies return the same output.

For maximality, the argument assumes that the rectangle is the only available joint information. If each task interval intersects its acceptance set, a choice of one value from each intersection defines a joint vector. This joint vector belongs to $\mathcal U_j$ and is feasible. Rejecting that candidate therefore risks removing a feasible vector consistent with the available information. Outer rejection is the strongest universally safe candidate-wise rule that retains every feasible candidate.

This argument requires a Cartesian-product uncertainty set, not probabilistic independence. Additional coupling information potentially supports stronger pruning. A policy that preserves only the first output also permits skipping later candidates once an earlier feasible output is known, but it pursues a different objective.

At $A_{c,j}=0$, the interval is a singleton and inner and outer eligibility coincide for that task. At $\epsilon_p=0$, outer absolute preservation requires $|P_{p,j}|\leq A_{p,j}$, while inner preservation requires $P_{p,j}=A_{p,j}=0$. The exact weak inequalities retain equalities. A positive target floor rejects a known unchanged baseline with $P=A=0$. If the bounds are probabilistic, the result holds on their joint coverage event over the frozen finite pool; marginal coverage does not supply that event.

The envelope concerns the exact finite validation objective used by the comparator. A search prediction requires a bound on
\begin{equation}
 \Delta_c^{\rm val}-P_c^{\rm search}
 =(P_c^{\rm val}-P_c^{\rm search})
   +\mathcal R_c^{\rm val}.
\end{equation}
A Taylor bound on the final term alone does not cover the search-to-validation error. Preserving the validation output also says nothing about independent test eligibility.

The output guarantee assumes a nonbinding verification budget. Under a common cap, screening can reach a feasible candidate that direct checking cannot reach, so equal outputs or abstentions are not guaranteed. The check-count comparison excludes envelope construction and calibration.

\section{Implementation Details}
\label{app:implementation}\label{app:replication}

This section details the method implementation of Section~\ref{sec:setup} and the experimental setup of Section~\ref{sec:study}: loss and item allocation, spectral and gradient audits, the development pilot, the proposal protocol, execution, and cost accounting.

\subsection{Loss, Splits, and Item Allocation}
\label{app:loss-splits}

\begin{table*}[!tp]
\centering\small
\begin{tabular}{lll}
\toprule
Suffix within \texttt{model.layers.}$\ell$ & Weight shape (out, in) & Role\\
\midrule
\texttt{self\_attn.q\_proj}, \texttt{self\_attn.o\_proj} & $4096\times4096$ & Attention\\
\texttt{self\_attn.k\_proj}, \texttt{self\_attn.v\_proj} & $1024\times4096$ & Attention\\
\texttt{mlp.gate\_proj}, \texttt{mlp.up\_proj} & $14336\times4096$ & Feed-forward\\
\texttt{mlp.down\_proj} & $4096\times14336$ & Feed-forward\\
\bottomrule
\end{tabular}
\caption{Mistral-7B-Instruct-v0.3 projection inventory from official configuration, parameter names, and implementation \citep{mistral2024configuration,mistral2024weightindex,transformers2026mistral}. Layers are zero-indexed from 0 to 31. Only the declared interior-layer down projections are editable. Shapes describe architecture metadata, not finite-edit measurements.}
\label{tab:architecture}
\end{table*}

Official architecture metadata lists 7 projections in each of 32 decoder layers, giving 224 projection matrices \citep{mistral2024configuration,mistral2024weightindex}. Table~\ref{tab:architecture} records their dimensions; the preflight checks the module inventory against the pinned checkpoint before decomposition.

The editable set is $\{\texttt{model.layers.}\ell\texttt{.mlp.down\_proj}:\ell=2,\ldots,29\}$, with zero-based layer indices. Each weight is $4096\times14336$. The leading 256 singular components per matrix give a support of $28\times256=7168$ coordinates.

\paragraph{Loss normalization.} Before scoring, the preflight rejects any missing or mismatched module, basis, coordinate table, or example-ID manifest. The protocol pins dataset versions, prompts, tokenizer, answer masks, precision, and the loss implementation. On a fixed split of $n_c$ items, the objective is
\begin{equation}
\begin{aligned}
 L_c(\theta)&=\frac{1}{n_c}\sum_{r=1}^{n_c}\ell_r(\theta),\\
 \ell_r(\theta)&=-\frac{1}{|y_r|}\sum_{k=1}^{|y_r|}\log p_\theta(y_{r,k}\mid x_r,y_{r,<k}).
\end{aligned}
\end{equation}
The loss implementation masks prompt tokens, averages answer-token losses within each item, and then averages items. The protocol fixes EOS inclusion, truncation, empty-answer rejection, and padding masks. Gradient accumulation uses the same item weights: averaging unequal microbatch averages generally changes the objective. A dataset's task label does not specify these choices.

\paragraph{Basis and edit execution.} The protocol computes the SVD once in float32 and stores the backend, precision, basis checksum, reconstruction errors, and singular gaps around editable indices. It serializes the selected basis within repeated-value subspaces.

Before each mask, the execution procedure restores the base weights, applies the physical displacement, verifies the read-back, and then initializes any key/value cache. The intervention includes neither compensation nor retained-weight refitting. The audit reports reconstruction errors before and after conversion to inference precision.

The implementation check precedes full-deletion scoring. It compares automatic gradients with centered finite differences on 20 fixed sampled coordinates at amplitude \ensuremath{10^{-3}}, using the experiment's item weighting and frozen basis. The check records signed derivatives, finite-difference estimates, and absolute and relative discrepancies. It handles near-zero derivatives without division by zero. This check tests the derivative implementation, not linearity under full deletion.

The primary pool audit uses forward scoring. Optional explanatory measurements recompute split derivatives on frozen supports and evaluate full-displacement HVPs. When these measurements are included, the amplitude grid is $\{0,0.01,0.05,0.1,0.25,0.5,1\}$, with quadrature refinements of 16, 32, and 64 subintervals. The optional audit records disagreement between refinements. These measurements do not certify a path supremum, and a fitted error coefficient remains distinct from a Hessian bound.

\paragraph{Correctness readouts.} GSM8K uses a frozen answer-extraction policy. HumanEval and MBPP+ use one greedy completion with a pinned execution environment, sanitizer, generation limit, and timeout policy. MBPP+ v0.2.0, derived from MBPP \citep{austin2021mbpp} with EvalPlus test augmentation \citep{liu2023evalplus}, provides 378 external test tasks; the protocol pins its task-list digest and the EvalPlus harness revision.\footnote{MBPP+ release documentation: \url{https://github.com/evalplus/evalplus}.} Test scoring evaluates reference-code answer-only NLL for selected edits and the baseline only, outside calibration. ARC-C uses the frozen answer-choice evaluation. The protocol retains item-level baseline and edit outcomes for every benchmark to support paired comparisons.

\paragraph{Uncertainty.} The test-evaluation protocol uses 10000 paired item resamples for NLL and correctness differences, with marginal 95\% intervals and a declared familywise rule across target and preserved tasks. A support selected by multiple policies shares one interval record. The protocol specifies any additional multiplicity control across distinct supports before testing.

Exact frozen-pool survivor counts do not receive item-bootstrap intervals. Direction-level summaries describe the proposal family. Constructing the paired intervals requires item-level records beyond aggregate means or totals.

\paragraph{Numerical comparisons.} The execution export retains unrounded losses, energies, and predictions. Verification replay uses decimal arithmetic at 50-digit working precision and rounds comparison quantities to 12 decimal places, ties to even. The same rule applies to direct feasibility, coverage, and inner and outer thresholds, retaining equalities. The replay records decisions that change without rounding. This convention specifies the finite numerical comparator; the analysis uses the exact identities.

\label{app:deferred-protocol}

\begin{table}[!tp]
\centering\scriptsize
\setlength{\tabcolsep}{3pt}
\begin{tabular}{lrrr}
\toprule
Task & Search & Validation & Test\\
\midrule
GSM8K & 512 & 512 & 1319\\
ARC-C & 256 & 128 & 1172\\
HumanEval & 64 & 32 & 68\\
MBPP+ & 0 & 0 & 378\\
\bottomrule
\end{tabular}
\caption{Fixed item allocation for search, validation, and test. Example IDs and dataset revisions are frozen before search-gradient computation. MBPP+ provides an untouched external test readout.}
\label{tab:study-splits}
\end{table}

The basis procedure computes the thin SVD once in float32 and serializes it with reconstruction errors and singular-value gaps. Projecting item-averaged gradients from 832 search items onto this basis gives 21504 coefficients (Equation~\ref{eq:gamma}). The 3 objectives use accumulated microbatch backward passes. Before finite-deletion scoring, centered finite differences on 20 coordinates at amplitude \ensuremath{10^{-3}} check the implementation (Appendix~\ref{app:gradient-audit}).

\subsection{Basis-Only Spectrum and Energy Audit}
\label{app:basis-audit}

The audit separates architecture discovery, spectral measurement, and task evaluation. The spectral audit records matrix spectra and reconstruction diagnostics before the protocol selects energy strata. Task scoring then reuses the fixed basis. Table~\ref{tab:basis-only-audit} summarizes the spectrum and deletion-energy distribution for the declared support.

We define $E_{\rm full}=\sum_{m\in\calM}\norm{W_m}_F^2$ and $E_{\rm band}=\sum_{i\in\calI}e_i$. The band fraction is $E_{\rm band}/E_{\rm full}$. Reconstruction with the full thin basis tests numerical accuracy. Retaining only the leading searchable components has relative Frobenius tail error $\sqrt{1-E_{\rm band}/E_{\rm full}}$; this tail error differs from the full-SVD residual. The boundary gap is $(\sigma_{m,256}-\sigma_{m,257})/\sigma_{m,256}$. Repeated positive values give zero gap; a zero denominator gives an undefined gap.

For $w_i=e_i/E_{\rm band}$, define $N_{\rm eff,E}=1/\sum_iw_i^2$. Sampling exactly $B$ coordinates uniformly without replacement gives
\begin{equation}
\begin{aligned}
 \E[\bar E]&=B/N,\\
 \operatorname{Var}(\bar E)&=\frac{B(N-B)}{N(N-1)}
 \left(\frac{1}{N_{\rm eff,E}}-\frac{1}{N}\right).
\end{aligned}
\label{eq:uniform-energy-variance}
\end{equation}
Equal inclusion probabilities give the mean. Negative covariance between distinct coordinate indicators gives the variance of the weighted sum. The variance vanishes when all energies are equal. These moments determine neither pilot quantiles nor the validity of a Gaussian approximation; the recorded uniform masks supply the intervals in Appendix~\ref{app:real-generator}.

\begin{table*}[!tp]
\centering\small
\begin{minipage}[t]{0.49\textwidth}
\centering
\setlength{\tabcolsep}{3pt}
\begin{tabular}{lr}
\toprule
Spectrum field & Value\\
\midrule
$E_{\rm full}$ & \newnum{486000}\\
$E_{\rm band}$ & \newnum{117600}\\
Band energy fraction & \newnum{0.241975}\\
Matrix $\sigma_1$ range & \newnum{8.3}--\newnum{25.8}\\
Matrix $\sigma_{256}$ range & \newnum{2.6}--\newnum{4.5}\\
Gap minimum/median/maximum & \newnum{0.0002}/\newnum{0.0011}/\newnum{0.0038}\\
$N_{\rm eff,E}$ & \newnum{2000}\\
Full-SVD relative residual, max & $\newnum{4.7\times10^{-6}}$\\
Orthogonality residual, max & $\newnum{3.8\times10^{-6}}$\\
Leading-band relative tail error & \newnum{0.870646}\\
\bottomrule
\end{tabular}
\end{minipage}\hfill
\begin{minipage}[t]{0.49\textwidth}
\centering
\setlength{\tabcolsep}{3pt}
\begin{tabular}{lr}
\toprule
Uniform-mask field & Value\\
\midrule
Requested pilot masks & \newnum{16384}\\
Minimum/maximum energy & \newnum{0.126}/\newnum{0.162}\\
$Q_{.25}(\bar E)$ & \newnum{0.138}\\
$Q_{.50}(\bar E)$ & \newnum{0.144}\\
$Q_{.75}(\bar E)$ & \newnum{0.150}\\
Pilot energy mean & \newnum{0.1429}\\
Pilot energy standard deviation & \newnum{0.0067}\\
Exact uniform expectation & \newnum{0.142857}\\
SD conditional on $N_{\rm eff,E}$ & \newnum{0.006644}\\
\bottomrule
\end{tabular}
\end{minipage}
\caption{Spectral diagnostics (left) and uniform-mask deletion energy (right) for the primary support. Energy fraction, leading-band tail error, uniform expectation, and theoretical standard deviation are calculated from the reported inputs. Reconstruction residuals use the full thin SVD. The orthogonality residual is the larger of $\norm{U^\top U-I}_F/\sqrt{r_m}$ and $\norm{V^\top V-I}_F/\sqrt{r_m}$. Uniform-mask quantiles are reported separately from the variance calculation.}
\label{tab:basis-only-audit}
\end{table*}

\subsection{Search Saliency and Finite-Difference Audit}
\label{app:gradient-audit}

Search uses \newnum{512} GSM8K, \newnum{64} HumanEval, and \newnum{256} ARC-C items. The gradient procedure accumulates each task's item-averaged objective separately, with the checkpoint in evaluation mode and the singular basis fixed. It projects all \newnum{7168} coordinates for each task and retains signed coefficients. A positive coefficient predicts a loss increase under deletion; a negative coefficient predicts an improvement. Neither sign licenses interpreting an absolute sum as the response of uniform shrinkage.

For each task, the signed score sum is $T_c=\sum_i\gamma_{c,i}$, the absolute score sum is $A_c=\sum_i|\gamma_{c,i}|$, and the squared score sum is $M_{2,c}=\sum_i\gamma_{c,i}^2$. The effective magnitude count is $N_{\rm eff,c}=A_c^2/M_{2,c}$ when $M_{2,c}>0$. Table~\ref{tab:gradient-summary} reports these moments. A magnitude histogram, conditional-sign reference rank, or selected subset's prediction requires the complete coefficient arrays.

\begin{table*}[!tp]
\centering\small
\setlength{\tabcolsep}{4pt}
\begin{tabular}{lrrrrrr}
\toprule
Task & Positive & Negative & Signed $T_c$ & $A_c$ & $M_{2,c}$ & $N_{\rm eff,c}$\\
\midrule
GSM8K & \newnum{3742} & \newnum{3426} & \newnum{+0.0578} & \newnum{0.920} & \newnum{0.000384727273} & \newnum{2200}\\
HumanEval & \newnum{3526} & \newnum{3642} & \newnum{-0.0112} & \newnum{0.760} & \newnum{0.000313913043} & \newnum{1840}\\
ARC-C & \newnum{3806} & \newnum{3362} & \newnum{+0.0204} & \newnum{0.640} & \newnum{0.000173559322} & \newnum{2360}\\
\bottomrule
\end{tabular}
\caption{Search-gradient saliency across spectral coordinates. Each task has \newnum{7168} nonzero coefficients; positive and negative counts sum to this total. Signed totals and absolute magnitudes are reported separately. Effective counts use the absolute and squared score sums, subject to displayed rounding.}
\label{tab:gradient-summary}
\end{table*}

The check samples one fixed set of \newnum{20} coordinate IDs before evaluating finite differences and reuses it across the three objectives. For a coordinate direction $d_i=-A_i$, the check compares $\gamma_{c,i}$ with
\begin{equation}
 \begin{aligned}
 \widehat\gamma_{c,i}^{\rm FD}
 &=\frac{L_c(\theta+\delta d_i)-L_c(\theta-\delta d_i)}{2\delta},\\
 \delta&=\newnum{10^{-3}}.
 \end{aligned}
\end{equation}
Both endpoints use the same examples, masks, batching, and inference precision. The check accumulates item losses and their subtraction in float\newnum{64}; this does not eliminate forward-rounding error. It stores each signed pair and its absolute discrepancy. Relative disagreement applies only where $|\gamma_{c,i}|$ exceeds the declared near-zero threshold. The check excludes \newnum{2}, \newnum{2}, and \newnum{3} near-zero coordinates for GSM8K, HumanEval, and ARC-C, respectively.

\begin{figure*}[!tp]
\centering
\includegraphics[width=\textwidth]{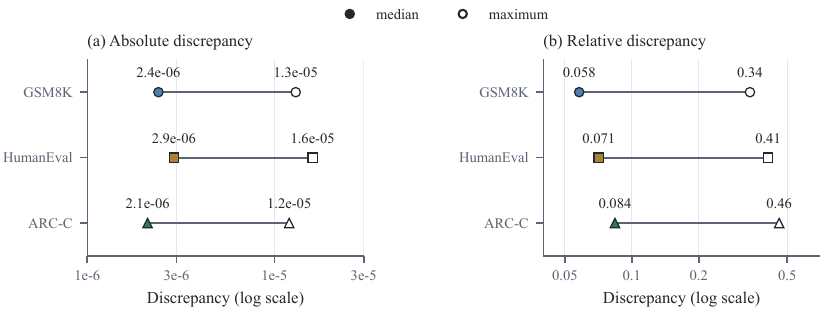}
\caption{Finite-difference discrepancies for \newnum{20} shared spectral coordinates per task at $\delta=\newnum{10^{-3}}$. Filled and hollow markers show median and maximum discrepancies; segments connect these summary values. Absolute discrepancy is $|\widehat\gamma^{\rm FD}-\gamma|$; relative discrepancy divides by $|\gamma|$ only when $|\gamma|>\newnum{10^{-5}}$. Both horizontal axes use log scales.}
\label{fig:finite-difference}
\end{figure*}

If the check fails, the protocol requires implementation correction before pool scoring and keeps task tolerances fixed. Under the shared-coordinate design, the two-sided check adds \newnum{33280} search-item scores; unedited losses can be reused. Figure~\ref{fig:finite-difference} summarizes the check.


\subsection{Development Pilot}
\label{app:dev-pilot}

Pool construction begins with the energy-only pilot in Appendix~\ref{app:real-generator}, which identifies energy ranges without evaluating task losses. A separate development pilot then generates 64 diagnostic masks under a provisional configuration and scores their finite validation losses. It records feasibility, error signs, and how often the target prediction exceeds the measured change.

An informative pilot has a feasible fraction between approximately 5\% and 60\%, errors of both signs, and at least one target loss change above its prediction. These criteria test whether the pool can expose screening behavior. If they fail, the development stage revises target floors, preserved tolerances, or energy allocation and records each change. The study stops if the development stage produces no informative configuration. The configuration log retains pilot outcomes, and final coverage summaries exclude them.

Support hashes exclude pilot masks from both final pools. The pilot reuses the declared validation items, so subsequent coverage is conditional on that corpus; the independent test split remains untouched. After the pilot, the protocol freezes thresholds, support, strata, generator settings, and loss policy. The protocol seals both the 256-mask calibration manifest and the 1024-mask evaluation manifest before scoring either pool.

\subsection{Outcome-Blind Proposal Protocol}
\label{app:real-generator}

Every proposal meets the exact coordinate budget. The first mask selects the largest target derivatives; later proposals cycle through target-focused scores, scores penalizing preserved-task magnitudes, and random scores, with deterministic swaps keeping energies within the declared ranges. Finite loss changes never enter proposal scores. The records retain duplicate supports, retries, pilot fallbacks, and achieved stratum counts; the procedure below gives the seed, hashing, and repair procedure.

\paragraph{Inputs and deterministic anchor.} This protocol generates coordinate masks independently of finite outcomes. The set $\calI$ contains the 7168 serialized coordinates in fixed matrix/index order. The protocol computes the search derivatives in Equation~\ref{eq:gamma} and defines
\begin{equation}
\begin{aligned}
 U&=\{i\in\calI:e_i>0\},\\
 D&=E(\calI)=\sum_{i\in\calI}e_i,\\
 \bar E(S)&=\sum_{i\in S}e_i/D,\\
 P_c(S)&=\sum_{i\in S}\gamma_{c,i}.
\end{aligned}
\end{equation}
The protocol requires finite inputs, $D>0$, and $B\leq|U|$. Every mask contains exactly $B$ coordinates. Reductions use float64 and sorted coordinate IDs; the protocol fixes the software version. The first mask $S_0$ selects the $B$ largest target scores, with smaller-ID ties, and records its nonpositive-score count. Its finite outcomes do not influence inclusion.

\paragraph{Seeds and support identities.} The protocol records its version, master seed 20261006, checkpoint digest, basis digest, search-example IDs, loss configuration, gradient-array digest, target, and budget in canonical sorted-key UTF-8 JSON. This complete metadata defines the cell identity.

The seed procedure derives each unsigned 64-bit PCG64 seed from the first 8 SHA-256 bytes of the canonical tuple (cell identity, purpose, slot, attempt), interpreted little-endian. The energy-only pilot uses purpose \texttt{energy-pilot}, slot -1, attempt 0; proposals use purpose \texttt{proposal} and their actual slot/attempt.

A mask hash includes protocol/checkpoint/basis identities, target, $B$, and sorted coordinate IDs. Family labels and outcomes are excluded, so duplicate supports cannot acquire new identities through relabeling.

\paragraph{Energy-only pilot and strata.} The energy-only pilot, which computes no task losses, requests
\begin{equation}
 M_* = \min\{16384,\tbinom{|U|}{B}\}
\end{equation}
uniform masks. The pilot draws $B$ coordinates without replacement, sorts them, and rejects duplicate hashes. It stops at $M_*$ distinct masks or 32$M_*$ draws and reports the achieved $M$.

The pilot sorts masks by $(\bar E,\text{hash})$ and splits them into four consecutive groups at positions $\lfloor sM/4\rfloor$, $s=0,\ldots,4$. Group $\mathcal Q_s$ supplies an energy interval $[L_s,U_s]$ from its extreme energies. The pilot reports touching or coincident intervals; energy ties do not establish distinct regimes. The protocol fixes
\begin{equation}
 \bar E_{\rm cap}=\max\{\bar E(S_0),\max_{S\in\rm pilot}\bar E(S)\}
       +10^{-12}
\end{equation}
before outcomes. The added term is numerical tolerance, not a loss reserve. If $M<4$ or a stratum is empty, generation stops before computing its extrema. Invalid inputs also stop generation. 

\paragraph{Proposal scores.} For slot $r\geq1$, the protocol defines family $f=(r-1)\bmod3$, stratum $s=(r-1)\bmod4$, and block $b=\lfloor(r-1)/12\rfloor$. The normalization is
\begin{equation}
 a_c=\max\!\left\{10^{-12},\sqrt{\frac{1}{|U|}\sum_{i\in U}\gamma_{c,i}^2}\right\}.
\end{equation}
The protocol cycles $\eta_b$ through $(0.25,0.5,1,2)$ and $\beta_b$ through $(0.25,1,4,16)$, indexed by $b\bmod4$. Each attempt draws independent standard Gumbel scores $z_i$ from its proposal seed. The generator clips inverse-transform uniform inputs to representable open-interval endpoints. The three score families are
\begin{equation}
\begin{aligned}
 q_i^{(0)}&=\gamma_{t,i}/a_t+\eta_bz_i,\\
 q_i^{(1)}&=\gamma_{t,i}/a_t-\beta_b\sum_{p\in P}|\gamma_{p,i}|/a_p+\eta_bz_i,\\
 q_i^{(2)}&=z_i.
\end{aligned}
\end{equation}
The preserved-magnitude penalty discourages large preserved-task scores but does not certify net preservation. Optimizing random scores also differs from uniform support sampling. Reported predictions always use the unperturbed $\gamma$.

\paragraph{Energy-constrained swaps.} After drawing scores, the same attempt generator chooses an initial mask uniformly from $\mathcal Q_s$. At each step, the generator forms up to 64 lowest-score included IDs and 64 highest-score excluded IDs, with smaller coordinate ID breaking both score-sort ties.

The generator considers their swaps $i\to j$ only when the score gain is positive, the new energy belongs to $[L_s-10^{-12},U_s+10^{-12}]$, and the cap holds. It chooses the largest $q_j-q_i$, breaking ties by $(i,j)$.

The generator recomputes the proposed mask energy in sorted order and rechecks both interval and cap before applying the swap; it discards a numerically invalid pair. It stops after 64 swaps or when none qualifies. No optimality guarantee is attached to this bounded search.

\paragraph{Retries and fallback.} The generator rejects duplicate final hashes and retries for attempts 0 through 127.

If retries are exhausted, the generator ranks unused pilot masks in that stratum by $\sum_{i\in S}q_i$ under the final attempt's scores and smaller-hash ties. It accepts the best as a recorded pilot fallback. If none remains, generation terminates and reports the failed slot and achieved count.

Each accepted entry stores support, hash, family, interval, seeds, attempts, swaps, fallback status, energy, all search predictions, and cardinality; the rejection log separates pilot and proposal rejections.

\paragraph{Final-pool allocation.} The first mask plus 1023 accepted slots requests 1024 evaluation masks, with 341 additions per family and stratum additions $(256,256,256,255)$. The protocol assigns the anchor to the smallest-index stratum containing its actual energy. If none contains it, the protocol flags it as outside the calibrated range and uses direct verification. This also resolves touching or coinciding stratum boundaries.

The calibration pool uses 256 slots without the anchor, giving 64 requests per stratum before retries or failure. Achieved counts, not nominal counts, enter every analysis.

\paragraph{Development and exclusion.} The protocol generates the loss-evaluated development masks with purpose \texttt{development}, slots 1 through 64, and the same family and energy-repair rules. It reserves $S_0$ before this stage and excludes it from development, so its mandatory evaluation inclusion remains compatible with pilot exclusion. It requires distinct development supports. This stage precedes the final freeze.

Development scoring evaluates finite validation losses on the declared 672 items and records achieved counts, the feasible fraction, taskwise error-sign counts, target-overprediction counts, and every development decision. Revisions to thresholds, energy strata, and proposal settings occur only here and require a configuration-change log.

The support, basis, and search gradients remain fixed during this development stage; changing them requires a new cell and a new anchor preflight.

The protocol excludes every attempted loss-evaluated pilot support from calibration and evaluation, even if a later configuration changes its family label. The exclusion rule uses a physical-support exclusion digest of checkpoint, basis, target, budget, and sorted coordinate IDs, independent of the versioned run identity. The protocol discloses reuse of the fixed validation corpus; untouched test items never enter the pilot.

\paragraph{Final freeze.} After the development decision, the protocol seals the support, basis, gradients, thresholds, strata, generation settings, and both final manifests. The evaluation purpose is \texttt{proposal}; calibration uses \texttt{calibration}. It maintains a global support-hash exclusion set for the development masks and accepted final masks.

The generator constructs evaluation masks first and then calibration masks, excluding evaluation hashes. The protocol computes no finite loss from either final pool before sealing both manifests. The execution record stores the timestamp and canonical byte digest of the actual seal and records digests and run identifiers only when that execution occurs.

\paragraph{Calibration and verification.} Radius coefficients use the separate requested pool of 256 calibration masks, generated with purpose \texttt{calibration}, starting at slot 1, and rejecting evaluation-pool hashes as well as calibration duplicates. The deterministic anchor is excluded from calibration. The calibration record reports the achieved count in each energy stratum. An empty calibration stratum or candidate outside calibrated energy intervals bypasses screening and goes to direct verification, in the retained target order.

For task $c$ and stratum $s$, calibration fits $\bar H_{c,s}$ as the empirical 0.90 quantile of $2|\Delta_c^{\rm val}-P_c^{\rm search}|/\bar E$ among its calibration masks, using the smallest order statistic reaching that quantile.

The calibration procedure uses the grid $\{0,0.25,0.5,0.75,1,1.25,1.5,2,3,4\}$. It chooses its smallest multiplier whose intervals jointly cover all relevant tasks on at least 95\% of calibration masks. If none qualifies, the procedure flags the calibration failure and uses direct verification as the declared fallback. It freezes coefficients and multiplier before consulting evaluation outcomes. This fitted envelope includes observed derivative transfer, but its calibration coverage is not a uniform certificate.

The evaluation reports held-out coverage and rejected feasible masks. Direct checking, inner-screen verification, and outer-screen verification use the same pool, the single decreasing search-target order, and common finite-check budgets $Q\in\{32,128,512\}$ in the large cell. The comparison reports curves against $Q$, including abstention after exhaustion; a binding budget permits different policy outputs.

Cost accounting separates proposal work, gradients, radius calibration, and finite verification. One validation mask requires 672 item scores under the declared allocation; batching determines forward-call counts. Actual calls and elapsed costs require synchronized measurement. Test outcomes follow selected-mask freezing. Test derivatives are optional explanatory measurements, not required for policy replay.

\paragraph{Normalized calibration records.} The calibration record stores each mask's actual energy and signed error before normalization. Equation~\ref{eq:calibration} takes a quantile of candidate-level error-to-energy ratios. Dividing an absolute-error quantile by a midpoint generally gives a different result because varying denominators can change the observation order. The operations agree when all energies are equal.

Combined-pool summaries and calibration-only coefficients use different records. A stratum coefficient is unavailable without calibration records.

\subsection{Execution Sequence and Work Ledger}
\label{app:execution-ledger}

The spectral audit precedes task scoring. The protocol loads the pinned checkpoint weights, verifies the declared modules, and decomposes the 28 matrices sequentially in float32. CPU execution requires sufficient memory for the weights and decomposition workspace; elapsed time depends on the numerical backend and requires direct measurement.

The editable tensors alone occupy 6.125 GiB in float32; separate unedited copies need the same space. The leading $U$, $S$, and $V$ arrays occupy 528510976 bytes, excluding full-spectrum records and serialization overhead. These sizes follow from the declared shapes and do not include the rest of the checkpoint or SVD workspace.

The protocol serializes the leading basis and spectrum once and reuses them for gradients and deletion. The architecture inventory alone supplies no singular values.

The execution procedure keeps the unedited projection tensors available during finite scoring. For each mask, it restores them, applies the frozen displacement, verifies the weights, and clears any cache or state left by the preceding model. Batched answer-only scoring uses the fixed item allocation. The timing record separates restoration, scoring, and synchronization. Mask and item counts alone do not determine forward-call counts or elapsed cost.

The ledger covers one development batch and both final pools. Calibration and evaluation require 860160 validation item scores; including development gives 903168. One reusable score record for the unedited validation baseline adds 672 scores. Each repeated pilot batch incurs additional work. The ledger accounts for energy-only proposals, search gradients, finite-difference checks, and optional path measurements separately from validation endpoints.

After the protocol freezes policy outputs, test scoring evaluates each distinct selected support once. The symbol $K$ denotes their number. With the baseline, the workload is $2937(K+1)$ reference-NLL item scores, $1765(K+1)$ greedy completions across GSM8K, HumanEval, and MBPP+, and $1172(K+1)$ ARC-C answer-choice evaluations. Policies selecting the same support reuse its record; abstention reuses the baseline.

Table~\ref{tab:cost-accounting} gives these calculated workloads for the baseline and three distinct primary-study edits. Elapsed cost requires separate timing records.

\subsection{Evaluation Cost Accounting}
\label{app:cost}

The workload includes basis construction, search gradients, energy-only proposals, the loss-evaluated development pilot, calibration, the full-pool audit, policy consultations, and selected-output testing. Each validation consultation scores \newnum{672} items for one mask. Calibration scores \newnum{256} masks, requiring \newnum{172032} item scores; the audit scores \newnum{1024}, requiring \newnum{688128}. These counts describe work rather than measured runtime. A \newnum{64}-mask development batch adds \newnum{43008} item scores, and all repeated or attempted loss-inspected pilot masks incur additional development cost. Table~\ref{tab:cost-accounting} gives the complete accounting.

\begin{table*}[!tp]
\centering\small
\setlength{\tabcolsep}{3pt}
\begin{tabular}{lrr}
\toprule
Stage & Units & Item evaluations\\
\midrule
SVD & \newnum{28} matrices & n/a\\
Energy-only proposal pilot & \newnum{16384} masks & n/a\\
Search-gradient accumulation & \newnum{3} objectives & \newnum{832}\\
Finite-difference endpoints & \newnum{40} coordinate edits & \newnum{33280}\\
Unedited validation baseline & \newnum{1} support & \newnum{672}\\
One development pilot batch & \newnum{64} masks & \newnum{43008}\\
Calibration & \newnum{256} masks & \newnum{172032}\\
Full-pool audit & \newnum{1024} masks & \newnum{688128}\\
Final-pool subtotal & \newnum{1280} masks & \newnum{860160}\\
Development plus final pools & \newnum{1344} masks & \newnum{903168}\\
With unedited validation baseline & \newnum{1345} supports & \newnum{903840}\\
\midrule
Selected-test reference NLL & \newnum{4} supports & \newnum{11748}\\
Greedy test completions & \newnum{4} supports & \newnum{7060}\\
ARC-C answer-choice correctness & \newnum{4} supports & \newnum{4688}\\
\bottomrule
\end{tabular}
\caption{Item-scoring workload for the primary study and selected edits. Subtotals overlap preceding rows and are not added again. Finite-difference scoring reuses \newnum{20} coordinate IDs across three objectives. Each validation mask uses \newnum{672} items. The baseline and edits A, B, and C each use \newnum{2937} reference-NLL items, \newnum{1765} greedy completions, and \newnum{1172} answer-choice evaluations. Counts describe item evaluations, not runtime; additional development batches, proposals, optional HVPs, weight restoration, forward calls, and timing require separate accounting.}
\label{tab:cost-accounting}
\end{table*}

The complete-pool audit makes it possible to measure prediction errors and identify rejected feasible masks. Replaying policies over those cached outcomes counts logical consultations, but replay time cannot establish end-to-end inference savings. A valid runtime comparison requires proposal and calibration setup as well as survivor verification under common batching and hardware. Its interpretation also depends on batch size, sequence lengths, precision, weight-restoration time, and synchronization.

\section{Additional Results}
\label{app:additional-results}\label{app:deferred}

This section extends the results of Section~\ref{sec:results} with error quantiles, the measured path residual, primary-study diagnostics, and selected-edit records.

\subsection{Prediction-Error Quantiles}
\label{app:error-quantiles}

Table~\ref{tab:error-quantiles} supplements Figure~\ref{fig:error-energy} with \newnum{0.95} error quantiles and calibration-only coefficients for every task and stratum.

\begin{table*}[!tp]
\centering\small
\setlength{\tabcolsep}{3pt}
\begin{tabular}{rlrrrr}
\toprule
Stratum & Task & $\bar E$ interval & $n$ & $Q_{.95}$ & $\bar H_{c,s}$\\
\midrule
\newnum{1} & GSM8K & \newnum{0.126}--\newnum{0.138} & \newnum{320} & \newnum{0.0230} & \newnum{0.278}\\
\newnum{1} & HumanEval & \newnum{0.126}--\newnum{0.138} & \newnum{320} & \newnum{0.0260} & \newnum{0.337}\\
\newnum{1} & ARC-C & \newnum{0.126}--\newnum{0.138} & \newnum{320} & \newnum{0.0190} & \newnum{0.211}\\
\newnum{2} & GSM8K & \newnum{0.138}--\newnum{0.144} & \newnum{320} & \newnum{0.0540} & \newnum{0.613}\\
\newnum{2} & HumanEval & \newnum{0.138}--\newnum{0.144} & \newnum{320} & \newnum{0.0300} & \newnum{0.324}\\
\newnum{2} & ARC-C & \newnum{0.138}--\newnum{0.144} & \newnum{320} & \newnum{0.0260} & \newnum{0.247}\\
\newnum{3} & GSM8K & \newnum{0.144}--\newnum{0.150} & \newnum{320} & \newnum{0.0570} & \newnum{0.623}\\
\newnum{3} & HumanEval & \newnum{0.144}--\newnum{0.150} & \newnum{320} & \newnum{0.0320} & \newnum{0.321}\\
\newnum{3} & ARC-C & \newnum{0.144}--\newnum{0.150} & \newnum{320} & \newnum{0.0270} & \newnum{0.276}\\
\newnum{4} & GSM8K & \newnum{0.150}--\newnum{0.162} & \newnum{320} & \newnum{0.0520} & \newnum{0.517}\\
\newnum{4} & HumanEval & \newnum{0.150}--\newnum{0.162} & \newnum{320} & \newnum{0.0350} & \newnum{0.302}\\
\newnum{4} & ARC-C & \newnum{0.150}--\newnum{0.162} & \newnum{320} & \newnum{0.0220} & \newnum{0.194}\\
\bottomrule
\end{tabular}
\caption{Additional prediction-error diagnostics by task and deletion-energy stratum. The $\newnum{0.95}$ quantiles summarize absolute search-to-validation NLL errors over \newnum{64} calibration and \newnum{256} evaluation masks per stratum. The $\bar H_{c,s}$ coefficients use calibration-only per-mask error-to-energy ratios. Medians and $\newnum{0.90}$ quantiles appear in Figure~\ref{fig:error-energy}.}
\label{tab:error-quantiles}
\end{table*}

Absolute-error quantiles measure discrepancy in NLL units; Equation~\ref{eq:calibration} measures error relative to each mask's energy. Normalization can reorder observations, so dividing an absolute-error quantile by a representative energy generally does not recover the calibrated coefficient. The conversion agrees when all energies are equal.

Uniform $B$-of-$N$ sampling gives $\E[\bar E]=B/N=\newnum{1}/\newnum{7}$ regardless of the singular energies, but this mean does not determine the energy strata. Appendix~\ref{app:basis-audit} reports the spectrum, band energy, reconstruction diagnostics, and uniform-mask energy distribution. These describe the geometry of deletion; Section~\ref{sec:results} measures its loss-prediction and screening consequences.

The strata compare different deletion energies at the same coordinate budget. Their interpretation also depends on achieved counts, within-stratum energy distributions, and proposal families. Absolute error and normalized error answer different questions: the median error can rise while the fitted coefficient falls. Understanding the observed trend requires both quantities.

Error signs indicate how a discrepancy can affect screening. Consistent target overprediction can make a screen appear safe despite invalid radii; underpredicted preserved-task changes can admit harmful edits. Positive and negative error counts reveal whether either pattern dominates the evaluated family. A negative preserved loss change improves reference likelihood but can still violate an absolute-stability constraint.

\subsection{Energy Scaling and Component Diagnostics}
\label{app:energy-scaling}

To describe how absolute error varies with normalized energy, we use the log-linear model
\begin{equation}
 \log |r_{c,j}|=a_c+b_c\log(\bar E_j/\bar E_0)+u_{c,j}
\label{eq:scaling}
\end{equation}
on nonzero errors, recording zero errors separately. Table~\ref{tab:scaling} fits the plotted stratum medians by equal-weight log-log least squares. This aggregate fit describes the displayed trend. Assessing zero errors, weighting choices, and energy overlap within proposal families requires additional candidate-level analysis.

The fit is descriptive because the selector does not randomize energy independently of other edit properties. Proposal family, layer allocation, coordinate magnitudes, and target scores can all vary with energy. Clarifying these effects requires comparisons within proposal families and checks of their energy overlap. A trend in the frozen pool does not establish a universal scaling law.

The two error components can be examined on representative masks chosen by an outcome-blind energy rule. This optional analysis recomputes their validation directional derivatives and subtracts them from the finite validation changes to obtain same-split remainders. Optional full-direction Hessian-vector products (HVPs) compare base-point curvature with the integrated path term. Full-pool forward scoring measures the combined error; these additional measurements explain its components and incur separate costs.

\subsection{Measured Residual along the Edit Path}
\label{app:measured-path}

Let $S$ be a frozen selected set. Its test-loss prediction and measured residual are
\begin{equation}
\begin{aligned}
 P_c(S,\alpha)&=\alpha\sum_{i\in S}\widehat\gamma_{c,i}^{\rm test},\\
 \widehat{\mathcal R}_c(S,\alpha)&=\widehat\Delta_c(S,\alpha)-P_c(S,\alpha).
\end{aligned}
\label{eq:measured-path}
\end{equation}
The measured residual uses the empirical test loss change and the derivative on that same split. Checking the integral representation requires independent Hessian-vector products along the edit path. A uniform certificate requires bounds for every eligible candidate, beyond the measured paths.

\subsection{Primary-Study Diagnostics}
\label{app:measurement-panels}

\begin{figure}[!tp]
\centering
\includegraphics[width=\columnwidth]{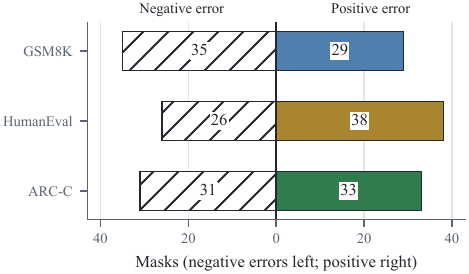}
\caption{Negative errors (hatched, left) and positive errors (solid, right) in the \newnum{64}-mask development pilot. Error is validation change minus search prediction; zero-error counts are \newnum{0} for all tasks. Of the masks, \newnum{17} are feasible (\newnum{26.6}\%) at target floor \newnum{0.12} and preservation tolerance \newnum{0.08}.}
\label{fig:pilot-error-signs}
\end{figure}

These diagnostics connect the development stage to the final screening and test outcomes. The primary study uses an arithmetic floor of \newnum{0.12} and preserved tolerances of \newnum{0.08}. Development masks are separate from the final \newnum{1280} calibration and evaluation masks. Figure~\ref{fig:pilot-error-signs} reports pilot feasibility and error signs; Table~\ref{tab:marginal-coverage} gives taskwise coverage; Figure~\ref{fig:coverage-retention}a gives joint coverage. Table~\ref{tab:scaling} gives the descriptive energy fit, and Table~\ref{tab:cost-accounting} accounts for the scoring workload.

\begin{table*}[!tp]
\begin{minipage}[t]{0.48\textwidth}


\centering\small
\setlength{\tabcolsep}{3pt}
\begin{tabular}{lrr}
\toprule
Task & Calibration & Evaluation\\
\midrule
GSM8K & \newnum{253} & \newnum{990}\\
HumanEval & \newnum{251} & \newnum{980}\\
ARC-C & \newnum{252} & \newnum{991}\\
\bottomrule
\end{tabular}
\caption{Taskwise interval coverage at the selected multiplier, \newnum{1.25}. Calibration and evaluation denominators are \newnum{256} and \newnum{1024}, respectively. Figure~\ref{fig:coverage-retention}a gives joint coverage.}
\label{tab:marginal-coverage}

\end{minipage}\hfill\begin{minipage}[t]{0.48\textwidth}

\centering\small
\setlength{\tabcolsep}{3pt}
\begin{tabular}{lrrr}
\toprule
Task & Scale $\exp(a_c)$ & Slope $b_c$ & Log-space $R^2$\\
\midrule
GSM8K & \newnum{0.010796} & \newnum{7.182} & \newnum{0.982}\\
HumanEval & \newnum{0.008528} & \newnum{5.145} & \newnum{0.994}\\
ARC-C & \newnum{0.006444} & \newnum{4.151} & \newnum{0.987}\\
\bottomrule
\end{tabular}
\caption{Descriptive log-log fits to the \newnum{4} stratum medians and energy midpoints in Figure~\ref{fig:error-energy}. Equal-weight least squares uses reference energy \newnum{0.141}. The fitted scale is the curve value at this reference, which need not equal the corresponding median. Coefficients describe the aggregate points and do not establish a mask-level scaling law.}
\label{tab:scaling}

\end{minipage}
\end{table*}

\subsection{Selected-Edit Validation and Test Records}
\label{app:readout-records}

A, B, and C each identify one selected-edit record; the labels are not support hashes. All three edits pass finite validation, but B misses the target floor on test and C exceeds the test code tolerance. These are validation-to-test differences, distinct from validation-envelope coverage.

Figure~\ref{fig:val-test} gives search predictions and finite validation and test NLL changes. Table~\ref{tab:output-nll-records} gives candidate-order positions and deletion energies, and Figure~\ref{fig:primary-correctness} gives correctness drops. The Mistral baseline NLL is $(\newnum{0.514},\newnum{0.683},\newnum{0.728},\newnum{0.756})$ in GSM8K, HumanEval, ARC-C, MBPP+ order. Adding the changes in Figure~\ref{fig:val-test} and Table~\ref{tab:output-nll-records} to these baselines gives the absolute endpoint NLL.

Baseline correctness counts are \newnum{900}/\newnum{1319} for GSM8K, \newnum{34}/\newnum{68} for HumanEval, \newnum{780}/\newnum{1172} for ARC-C, and \newnum{240}/\newnum{378} for MBPP+. Correctness drops use unrounded counts and are rounded only for display.

\begin{table}[!tp]
\centering\small
\setlength{\tabcolsep}{3pt}
\begin{tabular}{lrrr}
\toprule
Edit & Order position & $\bar E$ & MBPP+ $\Delta$ NLL\\
\midrule
A & \newnum{63} & \newnum{0.156} & \newnum{0.047}\\
B & \newnum{91} & \newnum{0.141} & \textbf{\newnum{0.039}}\\
C & \newnum{146} & \newnum{0.132} & \newnum{0.083}\\
\bottomrule
\end{tabular}
\caption{Selected-edit order positions, normalized deletion energies, and external MBPP+ test NLL changes. Positions refer to the full target-prediction order. MBPP+ NLL is outside radius calibration. Figure~\ref{fig:val-test} gives the three calibrated-task NLL changes, and Figure~\ref{fig:primary-correctness} gives correctness drops.}
\label{tab:output-nll-records}
\end{table}
\section{Checkpoint and Target-Role Extension}
\label{app:cross-model}

This section reports the checkpoint and target-role extension summarized in Section~\ref{sec:readouts}; it varies checkpoint, target role, and coordinate budget. With code as the target, HumanEval supplies differentiable search, calibration, and NLL selection; MBPP+ remains the primary external correctness readout. Arithmetic and ARC-C become preserved tasks.

Target floors are \newnum{0.12} for arithmetic and \newnum{0.18} for code. Preserved tolerances are \newnum{0.04} at budget \newnum{256} and \newnum{0.08} at \newnum{1024}. These task-specific thresholds do not imply equal capability loss across tasks, and their selection remains separate from test evaluation.

Table~\ref{tab:cross-cell} reports calibration multipliers and joint coverage for the additional configurations; Figure~\ref{fig:coverage-retention}d reports retention for all configurations. Mistral denotes Mistral-7B-Instruct-v0.3 \citep{jiang2023mistral,mistral2024modelcard} and Llama denotes Llama-3.1-8B-Instruct \citep{dubey2024llama3,meta2024llama31instruct}. In the Target column, A means arithmetic and C means code; these labels have a different role from the primary-study edit IDs.

\begin{table}[!tbp]
\centering\small
\setlength{\tabcolsep}{3pt}
\begin{tabular}{llrrrr}
\toprule
Model & Target & $B$ & $\kappa$ & Cal. joint & Eval. joint\\
\midrule
Mistral & A & \newnum{256} & \newnum{1.25} & \newnum{248} & \newnum{978}\\
Mistral & C & \newnum{256} & \newnum{1.50} & \newnum{249} & \newnum{982}\\
Mistral & C & \newnum{1024} & \newnum{2.00} & \newnum{247} & \newnum{958}\\
Llama & A & \newnum{256} & \newnum{1.25} & \newnum{246} & \newnum{974}\\
Llama & A & \newnum{1024} & \newnum{1.50} & \newnum{248} & \newnum{951}\\
Llama & C & \newnum{256} & \newnum{1.25} & \newnum{249} & \newnum{986}\\
Llama & C & \newnum{1024} & \newnum{2.00} & \newnum{247} & \newnum{962}\\
\bottomrule
\end{tabular}
\caption{Calibration multipliers and joint interval coverage for the additional configurations. Each uses \newnum{256} calibration and \newnum{1024} evaluation masks. Primary Mistral/arithmetic/\newnum{1024} coverage appears in Figure~\ref{fig:coverage-retention}a; panel (d) shows feasible-candidate retention for all configurations.}
\label{tab:cross-cell}
\end{table}

Each configuration uses \newnum{256} calibration masks and \newnum{1024} evaluation masks with checkpoint-specific support, gradients, calibration, and pilot. Coefficient transfer requires a separately declared audit. Figure~\ref{fig:budget-checks} compares verification under the common check caps.

\subsection{Capped Verification across Configurations}
\label{app:cross-verification}

\begin{figure}[!tp]
\centering
\includegraphics[width=\columnwidth]{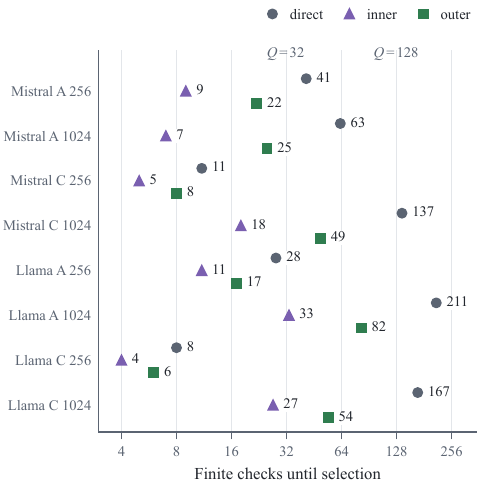}
\caption{Verification checks until selection at $Q=\newnum{512}$, where all policies select an edit. Circles, triangles, and squares denote direct verification, inner screening, and outer screening; adjacent numbers give check counts. Labels identify checkpoint, target (A: arithmetic; C: code), and coordinate budget $B$. Dashed lines mark $Q=\newnum{32}$ and $Q=\newnum{128}$; counts above a cap imply abstention under that cap. Direct and outer verification select the first direct-feasible support D; inner verification selects I. Verification counts exclude development, calibration, and full-pool auditing.}
\label{fig:budget-checks}
\end{figure}

Section~\ref{sec:verification} reports feasible-candidate retention, and Section~\ref{sec:budget-verification} summarizes the capped selection outcomes. Figure~\ref{fig:budget-checks} gives the exact verification counts for each policy and configuration.

\subsection{Selected Outputs across Checkpoints and Targets}

\begin{figure*}[!t]
\centering
\includegraphics[width=\textwidth]{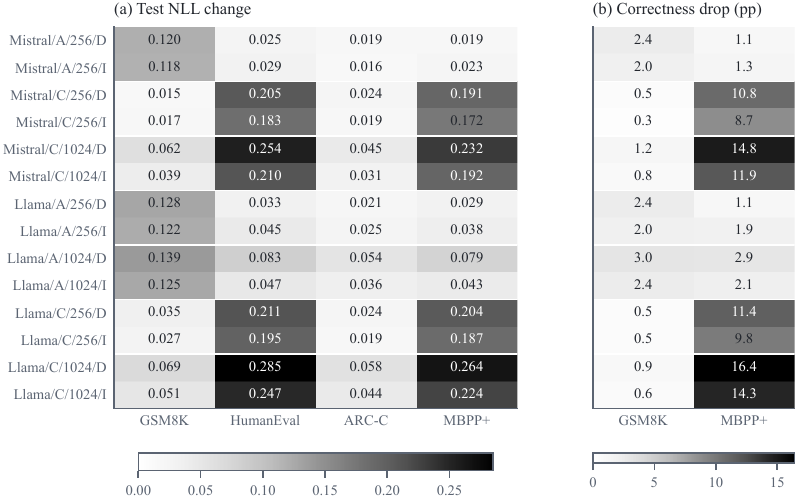}
\caption{Test NLL changes (a) and correctness drops in percentage points (b) for selected outputs in the additional configurations. Row labels identify checkpoint, target (A: arithmetic; C: code), coordinate budget, and output D or I. Positive values denote deterioration. Darker cells indicate larger changes within each panel; the panels use separate scales. Primary outputs appear in Figures~\ref{fig:val-test} and \ref{fig:primary-correctness}.}
\label{fig:extension-test}
\end{figure*}

Figure~\ref{fig:extension-test} reports test NLL changes and GSM8K and MBPP+ correctness drops for the additional configurations. Table~\ref{tab:extension-correct} gives supplementary HumanEval and ARC-C correctness counts. D denotes the first validation-feasible support in the direct order; I denotes the first validation-feasible inner survivor. For these configurations, direct and outer selections reuse D, and inner selections reuse I. The primary D and I outputs correspond to A and C, respectively; their results appear in Figures~\ref{fig:val-test} and \ref{fig:primary-correctness} and Table~\ref{tab:output-nll-records}.

\begin{table}[!tbp]
\centering\small
\setlength{\tabcolsep}{3pt}
\begin{tabular}{llrr}
\toprule
Cell & Output & HumanEval & ARC-C\\
\midrule
Mistral/A/\newnum{256} & D & \newnum{33} & \newnum{778}\\
Mistral/A/\newnum{256} & I & \newnum{33} & \newnum{779}\\
Mistral/C/\newnum{256} & D & \newnum{25} & \newnum{777}\\
Mistral/C/\newnum{256} & I & \newnum{27} & \newnum{778}\\
Mistral/C/\newnum{1024} & D & \newnum{22} & \newnum{768}\\
Mistral/C/\newnum{1024} & I & \newnum{24} & \newnum{775}\\
Llama/A/\newnum{256} & D & \newnum{37} & \newnum{854}\\
Llama/A/\newnum{256} & I & \newnum{36} & \newnum{851}\\
Llama/A/\newnum{1024} & D & \newnum{35} & \newnum{850}\\
Llama/A/\newnum{1024} & I & \newnum{36} & \newnum{853}\\
Llama/C/\newnum{256} & D & \newnum{29} & \newnum{854}\\
Llama/C/\newnum{256} & I & \newnum{31} & \newnum{855}\\
Llama/C/\newnum{1024} & D & \newnum{26} & \newnum{847}\\
Llama/C/\newnum{1024} & I & \newnum{28} & \newnum{852}\\
\bottomrule
\end{tabular}
\caption{Supplementary test correctness counts for the additional extension outputs. Denominators are \newnum{68} for HumanEval and \newnum{1172} for ARC-C. Figure~\ref{fig:extension-test} gives GSM8K and MBPP+ correctness drops; baseline counts are reported in the text. Aggregate counts alone do not supply paired confidence intervals.}
\label{tab:extension-correct}
\end{table}

The zero-radius primary output B appears in Figures~\ref{fig:val-test} and \ref{fig:primary-correctness}. No-edit outputs reuse the corresponding baseline and miss the positive target floor.

The Llama baseline NLL is $(\newnum{0.441},\newnum{0.612},\newnum{0.657},\newnum{0.674})$ in GSM8K, HumanEval, ARC-C, MBPP+ order; the Mistral baseline is in Appendix~\ref{app:readout-records}. Adding each reported change to its baseline gives the absolute endpoint NLL.

Baseline Llama correctness counts are \newnum{1000}/\newnum{1319}, \newnum{38}/\newnum{68}, \newnum{858}/\newnum{1172}, and \newnum{268}/\newnum{378} in the same benchmark order.

Interpretation of task-specific NLL changes includes GSM8K and MBPP+ correctness, abstentions, and both preservation criteria. The selected extension outputs all have nonnegative preserved-task NLL changes, so absolute and harm-only test decisions coincide here. An improvement exceeding the absolute tolerance produces different decisions under the two criteria. Comparisons between arithmetic and code remain conditional on their thresholds and readouts. Verification counts do not establish runtime savings.

For every code-target configuration, D produces both a larger HumanEval test NLL increase and a larger MBPP+ correctness drop than I, while both outputs keep the arithmetic and ARC-C NLL changes within the declared preservation tolerances (Figure~\ref{fig:extension-test}). The arithmetic-target configurations show different preservation outcomes: in the primary Mistral/\newnum{1024} study, D meets the test code tolerance and I exceeds it; in Llama/arithmetic/\newnum{1024}, D exceeds the same tolerance with a HumanEval NLL change of \newnum{0.083}, while I remains within it at \newnum{0.047}. Both Llama outputs exceed the arithmetic target floor. These fixed-output comparisons show why validation selection and test preservation require separate assessment, even when direct and outer verification return the same support.

\end{document}